\documentclass[10pt,journal,compsoc]{IEEEtran}

\ifCLASSOPTIONcompsoc
  \usepackage[nocompress]{cite}
\else
  \usepackage{cite}
\fi

\usepackage{times}
\usepackage{graphicx}
\usepackage{amsmath}
\usepackage{amsthm}
\usepackage{amssymb}
\usepackage{booktabs}
\usepackage{multirow}

\usepackage{algorithm}
\usepackage{algpseudocode}

\usepackage{subcaption}
\usepackage{adjustbox}

\usepackage{pifont}%
\newcommand{\cmark}{\ding{51}}%
\newcommand{\xmark}{\ding{55}}%

\usepackage[pagebackref=true,breaklinks=true,colorlinks,bookmarks=false]{hyperref}
\usepackage[capitalise, noabbrev]{cleveref}
\usepackage{orcidlink}

\newcommand*{\R}{\mathbb{R}}
\newcommand*{\X}{\mathcal{X}}

\newcommand*{\N}{\mathbb{N}}
\newcommand*{\Z}{\mathcal{Z}}

\newcommand*{\pr}{\mathbb{P}}
\newcommand*{\bfx}{\mathbf{x}}
\newcommand*{\bfz}{\mathbf{z}}

\DeclareMathOperator*{\ex}{\mathbb{E}}

\newcommand{\etal}{\textit{et al}.}
\newcommand{\ie}{\textit{i}.\textit{e}.}

\begin{document}
\title{Leveraging Diffusion For Strong and High Quality Face Morphing Attacks}

\author{Zander~W.~Blasingame\orcidlink{0000-0002-9508-8425}~
         and~Chen~Liu\orcidlink{0000-0003-1558-6836},~\IEEEmembership{Senior Member,~IEEE}%
\thanks{Manuscript received 23 February 2023; revised 8 June 2023; accepted 17
December 2023. This work was supported by the Center for Identification
Technology Research and National Science Foundation under Grant No. 1650503.
This article was recommended for publication by Associate Editor S. Lyu
upon evaluation of the reviewers’ comments. (Corresponding author:
Zander W. Blasingame.)
The authors are with the Department of Electrical and Computer
Engineering, Clarkson University, Potsdam, NY 13699 USA (e-mail:
blasinzw@clarkson.edu; cliu@clarkson.edu).
Digital Object Identifier 10.1109/TBIOM.2024.3349857}}

\IEEEpubid{2637-6047~\copyright~2024 IEEE. Personal use is permitted, but republication/redistribution requires IEEE permission.}

\IEEEtitleabstractindextext{%
\begin{abstract}

Face morphing attacks seek to deceive a Face Recognition (FR) system by presenting a morphed image consisting of the biometric qualities from two different identities with the aim of triggering a false acceptance with one of the two identities, thereby presenting a significant threat to biometric systems.
The success of a morphing attack is dependent on the ability of the morphed image to represent the biometric characteristics of both identities that were used to create the image.
We present a novel morphing attack that uses a Diffusion-based architecture to improve the visual fidelity of the image and the ability of the morphing attack to represent characteristics from both identities.
We demonstrate the effectiveness 
of the proposed attack by evaluating its visual fidelity via Fr\'echet Inception Distance (FID).
Also, extensive experiments are conducted to measure the vulnerability of FR systems to the proposed attack.
The ability of a morphing attack detector to detect the proposed attack is measured and compared against two state-of-the-art GAN-based morphing attacks along with two Landmark-based attacks.
Additionally, a novel metric to measure the relative strength between different morphing attacks is introduced and evaluated.
\end{abstract}

}

\maketitle

\IEEEdisplaynontitleabstractindextext

\IEEEpeerreviewmaketitle

\IEEEraisesectionheading{\section{Introduction}\label{sec:introduction}}

\IEEEPARstart{F}{ace} recognition (FR) systems have become one of the most common biometric modalities used for identity verification across a wide range of modern-day applications, from trivial tasks such as unlocking a smart phone to official businesses such as banking, e-commerce, and law enforcement.  
Unfortunately, while FR systems can often reach low false rejection and acceptance rates~\cite{frs-rates}, they are especially vulnerable to a new class of emerging attacks, known as the face morphing attack
~\cite{Ferrara2016,fraud_id,morphed_first,multi_image_attacks}.
The face morphing attack aims to compromise a fundamental property of biometric security, \ie, the one-to-one mapping from biometric data to the associated identity.
This compromise is achieved by creating a morphed face which contains biometric data of both identities in such a manner that presenting \textit{one} morphed image triggers a match with \textit{two} disjoint identities, violating the fundamental principle.

\begin{figure*}[t]
    \centering
    \begin{subfigure}{0.33\textwidth}
        \includegraphics[width=0.9\textwidth]{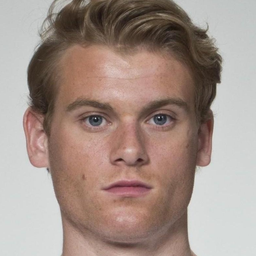}
        \caption{Identity $a$}
    \end{subfigure}%
    \begin{subfigure}{0.33\textwidth}
        \includegraphics[width=0.9\textwidth]{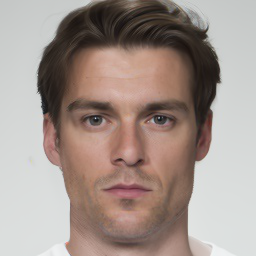}
        \caption{Morphed image}
    \end{subfigure}%
    \begin{subfigure}{0.33\textwidth}
        \includegraphics[width=0.9\textwidth]{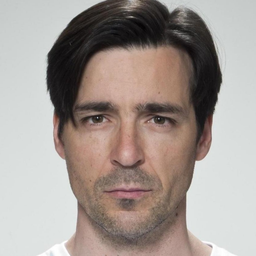}
        \caption{Identity $b$}
    \end{subfigure}
    \caption{Example of the proposed Diffusion-based morphing attack. Samples are from FRLL dataset.}
    \label{fig:morph_ex}
\end{figure*}

This poses a significant threat towards FR systems, especially in application cases such as e-passports and border access.
Notably the e-passport scenario, wherein the applicant submits a passport photo either in digital or printed form, is especially vulnerable to face morphing attack. This is particularly relevant for countries such as New Zealand, Estonia, and Ireland, where e-passports are used for both issuance and renewal of documents~\cite{Raghavendra2017FaceMV}.
In 2018 a German activist was reported to have received a German passport with a photo of his face morphed with an Italian politician~\cite{germany_morph_fraud}. Moreover, digital morphs can be easily generated hence offer a low-cost attack in the digital domain. It is not uncommon for a digital image submitted to a document submission portal to not have its authenticity verified by a human agent~\cite{mor-face-det-deep-resid}.
Critically, an adversary who is blacklisted from accessing a certain system can create a morph with a non-blacklisted individual to gain access.

Due to the severity of face morphing attacks, an abundance of algorithms have been developed to identify these attacks
~\cite{Ferrara2016,on-vuln,prnu,Blasingame2021LeveragingAL}.
Methods for Morphing Attack Detection (MAD) can be broadly characterized into two classes based on the manner in which they obtain the features used for detection, \ie, handcrafted features or deep features.
Handcrafted features are used in the so-called \textit{classical} algorithms which seek to find evidence of the morphing attack in the pixel domain, whether that is evidence of degradation in image quality~\cite{prnu}, residual noise left from the morphing attack~\cite{survery}, or local geometric features such as Local Binary Patterns (LBP)~\cite{Ojala1996ACS}, Binarized Statisical Image Feature (BSIF)~\cite{bsif}, and Local Phase Quantitization (LPQ)~\cite{morphed_first}.
Conversely, deep features are used with \textit{deep learning}-based algorithms.
These features are extracted by a deep Convolutional Neural Network (CNN), often a pre-trained network such as ResNet150~\cite{vgg19-is-best,mor-face-det-deep-resid}.
Generally, the best success has been found using deep CNN-based features in comparison to handcrafted features on both digital and print-scan face data~\cite{deep-best, NISTFRVT}.

Comparatively, there has been less research on face morphing attack algorithms.
Similar to the two classes of MAD algorithms, there exist two broad classes of face morphing attacks: Landmark-based attacks and deep learning-based attacks.
Landmark-based morphing attacks use local features to create the morphed image by warping and aligning the landmarks within each face then create a morphed face by pixel-wise compositing.
Landmark-based attacks have been shown to be effective against FR systems~\cite{sebastian_morphs}.
Recent work has enhanced the effectiveness of Landmark-based attacks by using adversarial perturbation~\cite{nasser_wavelet}.
In contrast, deep learning-based morphing attacks use a machine learning model to embed the original bona fide faces into a semantic representation which are then combined to produce a new representation that contains information from both identities.
This new representation is then used to generate a morphed face.

However, nearly all state-of-the-art deep learning based morph methods were based on the Generative Adversarial Network (GAN) framework~\cite{morgan, can_gan_beat_landmark}, with the primary difference being architectural improvements such as using the StyleGAN2 architecture~\cite{stylegan2} over the vanilla GAN architecture, or changes to the morph generation pipeline as seen in the Morphing through Identity Driven Prior GAN (MIPGAN)~\cite{mipgan}.
At the same time, there exists a handful of alternative state-of-the-art deep generative models
outside the GAN framework 
which offer their own advantages in terms of visual fidelity, semantic representation capabilities, and inference speed.

In particular, a class of generative models collectively known as ``Diffusion models''  haven been shown to possess high visual fidelity, even beating state-of-the-art GANs in visual fidelity~\cite{diff_beat_gan}, at the cost of increased inference time.
As visual fidelity and semantic representation abilities are far more important for the potency of a morphing attack than inference speed,
we present a novel methodology for generating strong face morphing attacks by leveraging Diffusion-based methods.
\cref{fig:morph_ex} shows an exemplary morphed image generated by the proposed attack constructed from two identities from the FRLL dataset~\cite{frll}.
We summarize the contributions of the proposed work as shown below:
\begin{itemize}
    \item We propose a novel method for generating morphed faces by using a Diffusion-based model which calculates twin embeddings to generate images of high visual fidelity.
    \item We evaluate our proposed attack against four other morphing attacks with extensive experiments assessing the vulnerability of three FR systems across three different datasets.
    \item The proposed morphing attack is evaluated via the Fr\'echet Inception Distance, a quantitative measure of visual fidelity.
    \item The proposed attack is further evaluated on its ability to evade detection from MAD algorithms trained against other morphing attacks.
    \item We introduce a novel metric to measure the strength of one morphing attack relative to another. 
    \item We present a small-scale study on the impact of pre-processing in the FR pipeline on the vulnerability of FR systems to morphing attacks.
    \item An exploration of different interpolation techniques on the proposed morphing attack is presented and evaluated. 
\end{itemize}

\section{Prior Work}

Several face morphing attacks have been developed by researchers with the morph generating process generally using face landmarks or deep learning.
In particular, we compare our proposed Diffusion-based attack against four state-of-the-art morphing attacks, two deep learning-based and two Landmark-based; namely, the OpenCV, FaceMorpher, StyleGAN2, and MIPGAN-II face morphing attacks.
These schemes represent different types of morphing attacks and provide a substantial baseline to measure the performance of the proposed Diffusion-based attack.
To the best of our knowledge all previously proposed deep learning-based attacks generate the morphed images via a type of GAN architecture~\cite{morgan,venkatesh2020gan,sebastian_morphs,mipgan}.

\subsection{Landmark-Based Morphing Attacks}
The FaceMorpher and OpenCV attacks were chosen as they are commonly used to represent Landmark-based attacks~\cite{sebastian_morphs,sebastian_gan_threaten,Blasingame2021LeveragingAL}.

\textbf{FaceMorpher} is an open-source algorithm that uses the STASM landmark detector~\cite{facemorpher,STASM}. From the landmarks on the images Delaunay triangles are formed, which are then warped and blended together.
The areas outside the landmarks are averaged, typically introducing strong artifacts in the neck and hair regions of the image~\cite{multe-scale-block-fusion}.

\textbf{OpenCV} morphing attack uses the open-source OpenCV library with a 68-point annotator from the Dlib library~\cite{dlib}.
The images and associated landmarks are used to form Delaunay triangles.
Then, in a similar manner to FaceMorpher, the landmarks are warped and blended.
In contrast to the approach of FaceMorpher, the areas outside the landmarks do not consist of an averaged image, but rather additional Delaunay triangles.
However, these morphs also exhibit strong artifacts outside the facial area due to the missing landmarks~\cite{multe-scale-block-fusion}.

\subsection{GAN-Based Morphing Attacks}
As mentioned earlier, prior deep learning-based attacks have used a GAN architecture for the morph generation process~\cite{sebastian_on_detection_of_ma_gan,sebastian_gan_threaten}.
GANs are a type of deep generative model which seeks to learn the sampling process for some data distribution $\pr_{data}$ on $\X$, \ie, given some simple distribution $Z \sim p(z)$ on $\Z$, the generator $G: \Z \to \X$ is to learn $G(Z) \sim \pr_{data}$.
A discriminator, sometimes called the critic, $D: \X \to [0,1]$ is trained adversarially against the generator in a minimax game described by
\begin{equation}
    \min_{G}\max_{D} \ex_{x \sim \pr_{data}} \log D(x) + \ex_{z \sim p(z)} \log (1 - (D \circ G)(z))
\end{equation}
where the discriminator attempts to get better at distinguishing synthetic samples from genuine samples, while the generator tries to get better at deceiving the discriminator.
For a GAN-based morphing attack, it becomes necessary for an encoding algorithm $E: \X \to \Z$ to exist, which can embed images in the latent space such that the inversion has low distortion $(G \circ E)(x) \approx x$.
The latent codes for two identities are then averaged to produce a new latent code representing the morphed face which is passed to the generator.
Notably, there exists a trade-off between the inversion distortion and editability of the latent embeddings~\cite{Roich2021PivotalTF}. 
Damer~\etal~proposed to use the GAN architecture for generating morphs by combining two latent codes encoded from two real identities to create a morphed code~\cite{morgan}.
This proposed attack, known as MorGAN, was based on a modification to the vanilla GAN architecture with the addition of an explicitly defined encoder architecture that was trained jointly with the generator via a modified adversarial loss formulation.
Since then the StyleGAN2~\cite{sebastian_morphs} and MIPGAN-II~\cite{mipgan} attacks have improved upon the MorGAN formulation by improving the GAN architecture, loss formulation, or encoding algorithm.

\textbf{StyleGAN2} offers a host of improvements over the standard GAN implementation that enables the architecture to achieve state-of-the-art image quality when generating high resolution images.
The StyleGAN2 model was pre-trained on the Flickr-Faces-HQ (FFHQ) dataset~\cite{stylegan}.
The faces were then cropped to possess the same landmark alignment as in the FFHQ dataset. Following the approach in~\cite{stylegan2}, the images, $x_a$, $x_b$, are embedded by optimizing an initial latent code through stochastic gradient descent, minimizing the perceptual loss between the generated image and target image.
After each embedded latent code, $z_a$, $z_b$, is found, a morphed latent code is created by linearly interpolating between the two, 
$z_{ab} = \textrm{lerp}(z_a, z_b; 0.5)$.
Lastly, the interpolated latent code is passed to the StyleGAN2 synthesis network to get the morphed image $x_{ab}$.
The StyleGAN2 morphs are strong when used with images containing a uniform background, which makes them especially powerful when used in conjunction with the Face Research Lab London (FRLL) dataset~\cite{frll}.

\textbf{MIPGAN-II} proposes an extension on StyleGAN2 by adding an optimization procedure for the latent vector used in creating the morphed image~\cite{mipgan}.
The StyleGAN2 portion of MIPGAN-II was pre-trained on the FFHQ dataset.
The two bona fide images are embedded into the latent space using the StyleGAN2 optimization procedure.
The latent code is initially constructed as 
$z_0 = \textrm{lerp}(z_a, z_b; 0.5)$.
For $n$ epochs the latent code is optimized to minimize a combination of perceptual loss, identity loss, identity difference loss, and Multi-Scale Structural Similarity loss, finding a fully optimized latent $z_n$.
The latent code $z_n$ is then passed to the StyleGAN2 synthesis network to create the morphed image.
As MIPGAN-II presents a refinement on StyleGAN2 for the application of morphing attack, it possesses similar advantages and disadvantages that StyleGAN2 morphs offer.

\begin{algorithm*}
    \caption{Diffusion Morphing Algorithm with DDIM scheduler where $\sigma_t = 0$.}
    \label{alg:morphing}
    \begin{algorithmic}
        \Require{The following components:
        \begin{enumerate}
            \item Bona fide images, $\bfx_0^{(a)}, \bfx_0^{(b)} \in \X$
            \item Noise prediction model $\boldsymbol\epsilon_\theta: \X \times \Z \times \N \to \X$
            \item Image space preprocessing function, $\xi : \X \times \X \to \X$
            \item Image space interpolation function, $\ell_\X : \X \times \X \times [0, 1] \to \X$
            \item Latent space interpolation function, $\ell_\Z : \Z \times \Z \times [0, 1] \to \Z$ \item Timing sub-schedule $\{\tau_i\}_{i=1}^N$ 
        \end{enumerate}}
        \Procedure{DiffusionMorph}{$\bfx_0^{(a)}, \bfx_0^{(b)}$, $\{\tau_i\}_{i=1}^N$}
        \State $\bfz_a \gets E(\bfx_0^{(a)})$ \Comment{Calculate semantic latent codes}
        \State $\bfz_b \gets E(\bfx_0^{(b)})$
        \State $\bfx_0^{(a)} \gets \xi(\bfx_0^{(a)}, \bfx_0^{(b)})$ \Comment{Preprocess images passed to stochastic encoder}
        \State $\bfx_0^{(b)} \gets \xi(\bfx_0^{(b)}, \bfx_0^{(a)})$
        \For {$i \gets 1, 2, \ldots, N$}
            \State $\bfx_{\tau_i}^{(a)} \gets \sqrt{\alpha_{\tau_i}} \bigg( \frac{\bfx_{\tau_{i-1}}^{(a)} - \sqrt{1 - \alpha_{\tau_{i-1}}} \cdot \boldsymbol\epsilon_\theta(\bfx_{\tau_{i-1}}^{(a)}, \bfz_a, \tau_{i-1})}{\sqrt{\alpha_{\tau_{i-1}}}}\bigg) + \sqrt{1 - \alpha_{\tau_i}}\boldsymbol\epsilon_\theta(\bfx_{\tau_{i-1}}^{(a)},\bfz_a, \tau_{i-1})$ \Comment{Forward pass of diffusion algorithm}
            \State $\bfx_{\tau_i}^{(b)} \gets \sqrt{\alpha_{\tau_i}} \bigg( \frac{\bfx_{\tau_{i-1}}^{(b)} - \sqrt{1 - \alpha_{\tau_{i-1}}} \cdot \boldsymbol\epsilon_\theta(\bfx_{\tau_{i-1}}^{(b)}, \bfz_b, \tau_{i-1})}{\sqrt{\alpha_{\tau_{i-1}}}}\bigg) + \sqrt{1 - \alpha_{\tau_i}}\boldsymbol\epsilon_\theta(\bfx_{\tau_{i-1}}^{(b)}, \bfz_b, \tau_{i-1})$
        \EndFor
        \State $\bfx_T^{(ab)} \gets \ell_\X (\bfx_T^{(a)}, \bfx_T^{(b)}; 0.5)$ \Comment{Stochastic code interpolation}
        \State $\bfz_{ab} \gets \ell_\Z (\bfz_{a}, \bfz_{b}; 0.5)$ \Comment{Semantic code interpolation}
        \For {$i \gets N, N - 1, \ldots 1$}
            \State $\bfx_{\tau_{i-1}}^{(ab)} \gets \sqrt{\alpha_{\tau_{i-1}}}\bigg(\frac{\bfx_{\tau_i}^{(ab)} - \sqrt{1 - \alpha_{\tau_i}}\boldsymbol\epsilon_\theta(\bfx_{\tau_i}^{(ab)}, \bfz_{ab}, \tau_i)}{\sqrt{\alpha_{\tau_i}}}\bigg)  + \sqrt{1 - \alpha_{\tau_{i-1}}} \cdot \boldsymbol\epsilon_\theta(\bfx_{\tau_i}^{(ab)}, \bfz_{ab}, \tau_i)$ \Comment{Diffusion generative process}
        \EndFor
        \State \textbf{return} $\bfx_0^{(ab)}$
        \EndProcedure
    \end{algorithmic}
\end{algorithm*}

\section{Diffusion-based Morphing Attack}

\begin{figure}[t]
    \centering
    \includegraphics[width=0.45\textwidth]{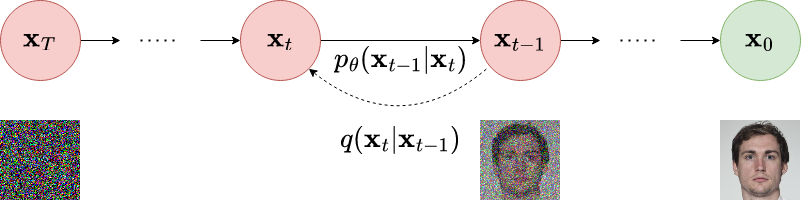}
    \caption{The forward and reverse Diffusion processes.}
    \label{fig:diffusion_process}
\end{figure}

Unlike GANs which learn the sampling process for the data distribution through adversarial training between the generator and the critic, diffusion-based and score-based generative models learn the data distribution via a denoising process through varying noise levels.
Diffusion-based models can achieve image fidelity superior to state-of-the-art generative models, matching even the acclaimed BigGAN-deep~\cite{biggandeep} model, while maintaining better coverage of the data distribution~\cite{diff_beat_gan}.
For these reasons we propose a morphing attack that uses Diffusion-based methods as the generative process.

\begin{figure*}
    \centering
    \begin{align}
        \label{eq:ddim_inference_dists}
        q_\sigma(\bfx_{t-1} \mid \bfx_t, \bfx_0) &= \mathcal{N} \bigg( \sqrt{\alpha_{t-1}}\bfx_0 + \sqrt{1 - \alpha_{t-1} - \sigma_t^2} \cdot \frac{\bfx_t - \sqrt{\alpha_t}\bfx_0}{\sqrt{1 - \alpha_t}}, \sigma_t^2 \mathbf{I} \bigg)\\
        \label{eq:decoder}
        \bfx_{t-1} &= \sqrt{\alpha_{t-1}}\underbrace{\bigg(\frac{\bfx_t - \sqrt{1 - \alpha_t}\boldsymbol\epsilon_\theta(\bfx_t, t)}{\sqrt{\alpha_t}}\bigg)}_{\textnormal{Predicted }\bfx_0} + \underbrace{\sqrt{1 - \alpha_{t-1} - \sigma_t^2} \cdot \boldsymbol\epsilon_\theta(\bfx_t, t)}_{\textnormal{Direction pointing to }\bfx_t} + \underbrace{\sigma_t \boldsymbol\epsilon_t}_{\textnormal{Random Noise}}\\
        \label{eq:decoder_accelerated}
        \bfx_{\tau_{i-1}} &= \sqrt{\alpha_{\tau_{i-1}}}\bigg(\frac{\bfx_{\tau_i} - \sqrt{1 - \alpha_{\tau_i}}\boldsymbol\epsilon_\theta(\bfx_{\tau_i}, \tau_i)}{\sqrt{\alpha_{\tau_i}}}\bigg)  + \sqrt{1 - \alpha_{\tau_{i-1}}} \cdot \boldsymbol\epsilon_\theta(\bfx_{\tau_i}, \tau_i)\\
        \label{eq:diffae_forward}
        \bfx_{\tau_i} &= \sqrt{\alpha_{\tau_i}} \bigg( \frac{\bfx_{\tau_{i-1}} - \sqrt{1 - \alpha_{\tau_{i-1}}} \cdot \boldsymbol\epsilon_\theta(\bfx_{\tau_{i-1}}, \bfz, \tau_{i-1})}{\sqrt{\alpha_{\tau_{i-1}}}}\bigg) + \sqrt{1 - \alpha_{\tau_i}}\boldsymbol\epsilon_\theta(\bfx_{\tau_{i-1}}, \bfz, \tau_{i-1})
    \end{align}
\end{figure*}

\begin{figure*}[t]
    \centering
    \includegraphics[width=\textwidth]{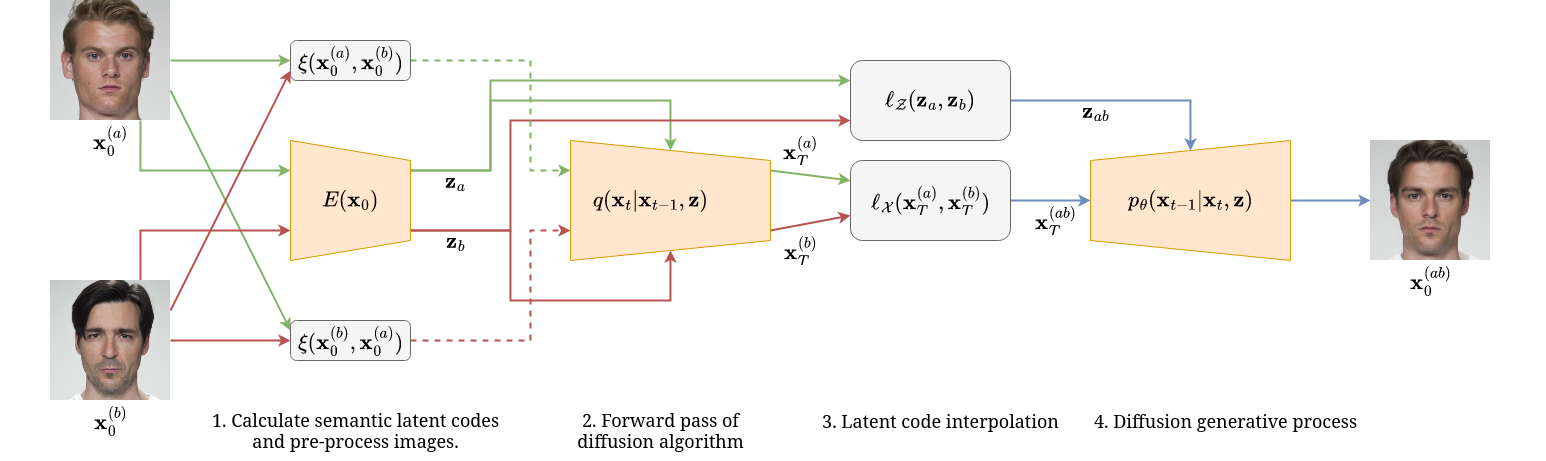}
    \caption{Proposed architecture for Diffusion-based morphs, where the green traces indicate variables associated with identity $a$, likewise red traces denote identity $b$, and blue traces for the morphed identity $ab$.}
    \label{fig:diffae}
\end{figure*}

\subsection{Diffusion Models}

Given data distribution $q(\bfx_0)$ on data space $\X$, the goal is to learn a model $p_\theta(\bfx_0)$ approximating $q(\bfx_0)$ which can be easily sampled.
Denoising Diffusion Probabilistic Models (DDPMs)~\cite{ddpm} are latent variable models of the form 
\begin{equation}
p_\theta(\bfx_0) = \int p_\theta(\bfx_{0:T})\; \textrm{d}\bfx_{1:T}    
\end{equation}
where $\{\bfx_t\}_{t=1}^T \in \X$ are latent variables and $T \in \N$. The reverse process is a Markov chain starting at 
$p_\theta(\bfx_T) = \mathcal{N}(\mathbf{0}, \mathbf{I})$, \ie, a normal distribution with mean vector $\mathbf{0}$ and variance $\mathbf{I}$ (the identity matrix),
with Gaussian transitions 
\begin{equation}
p_\theta(\bfx_{0:T}) = p_\theta(\bfx_0) \prod_{t=1}^T p_\theta(\bfx_{t-1}\mid \bfx_t)
\end{equation}
The diffusion (forward) process is fixed to a Markov chain that gradually adds Gaussian noise to the original sample $\bfx_0$ according to variance schedule $\{\beta_t\}_{t=1}^T$, such that 
\begin{equation}
q(\bfx_{1:T}, \bfx_0) = \prod_{t=1}^T q(\bfx_t\mid \bfx_{t-1})
\end{equation}
and
$q(\bfx_t|\bfx_{t-1}) = \mathcal{N}(\sqrt{1-\beta_t}\bfx_{t-1}, \beta_t\mathbf{I})$.
See \cref{fig:diffusion_process} for an illustration of this process.
The transition probability $p(\bfx_{t-1}|\bfx_t)$ is likely to be very complex, unless the gap between $t$ and $t-1$ is very small, \ie, $T \to \infty$. In this case, $p(\bfx_{t-1}|\bfx_t)$ can be modelled as $\mathcal{N}(\boldsymbol\mu_\theta(\bfx_t, t), \sigma_t)$, where $\boldsymbol\mu_\theta: \X \times \N \to \X$ is an estimator at step $t$ parameterized by $\theta$.
Ho~\etal~\cite{ddpm} proposes to use the following form:
\begin{equation}
    \boldsymbol\mu_\theta(\bfx_t, t) = \frac{1}{\sqrt{\alpha_t}}\bigg(\bfx_t - \frac{\beta_t}{\sqrt{1 - \bar\alpha_t}}\boldsymbol\epsilon_\theta(\bfx_t, t)\bigg)
\end{equation}
where 
$\alpha_t = 1 - \beta_t$,
$\bar\alpha_t = \prod_{s=1}^t \alpha_s$,
and $\boldsymbol\epsilon_\theta: \X \times \N \to \X$ is a function approximator parameterized by $\theta$, which learns to predict the noise added to $\bfx_0$ to get $\bfx_t$.
This is achieved by using a U-Net, a type of CNN consisting of several skip connections formed at each resolution size in the architecture to model $\boldsymbol\epsilon_\theta$~\cite{unet}.
Like variational autoencoder (VAE), the model is trained by optimizing the variational bound on the negative log likelihood.

Relaxing the constraint that the inference process has to be Markovian leads to another kind of diffusion model known as the Denoising Diffusion Implicit Model (DDIM) as proposed by Song \etal~\cite{song2021denoising}.
Consider a family $\mathcal{Q}$ of inference distributions indexed by a real vector $\sigma \in \R_{\geq 0}^T$
\begin{equation}
    q_\sigma(\bfx_{1:T} \mid \bfx_0) = q_\sigma(\bfx_T \mid \bfx_0) \prod_{t=2}^T q_\sigma(\bfx_{t-1} \mid \bfx_t, \bfx_0)
\end{equation}
where $q_\sigma(\bfx_T | \bfx_0) = \mathcal{N}(\sqrt{\alpha_T}\bfx_0, (1-\alpha_T)\mathbf{I})$ and for all $t > 1$~\cref{eq:ddim_inference_dists} holds.
This leads to the forward equation,~\cref{eq:decoder}.
Such a formulation allows for a deterministic generation of an image when $\sigma_t= 0$.

Another advantage of the DDIM framework is the ability to sample the forward trajectory more quickly by skipping timesteps.
Let $\{\tau_i\}_{i=1}^N$ be a monotonically increasing sequence of samples with $N < T$. By convention $\tau_N = T$ and an initial timestep $\tau_0 = 0$ is inserted. The inference equation can be reformulated into~\cref{eq:decoder_accelerated} where the parameter $N$ controls a trade-off between sample speed and visual fidelity.

Preechakul~\etal~\cite{diffae} proposed a Diffusion Autoencoder by employing a conditional DDIM.
The DDIM is conditioned on a semantic representation $\bfz \in \Z$ by modifying $\boldsymbol\epsilon_\theta^{(t)}$ to take $\bfz$ as an additional input.
An encoding network $E: \X \to \Z$ is introduced to learn the semantic latent code $\bfz = E(\bfx)$.
In this framework a twin pair of latent codes are created for a given image, $(\bfz, \bfx_T)$, the semantic and stochastic latent codes.
In theory, the semantic code controls semantic information like hair color, face shape, and other high level concepts, whereas the stochastic code controls stochastic variations in the image. Instead of sampling $\bfx_T \sim q_\sigma(\bfx_T \mid \bfx_0)$, Preechakul~\etal~proposed~\cref{eq:diffae_forward} as a deterministic forward process. This implementation necessitates that the semantic code $\bfz$ be generated first before creating the stochastic code.

\subsection{Proposed Morphing Algorithm}

We propose a novel process for the creation of morphed images by employing both the stochastic and semantic encoders.
In particular, let $x_a$, $x_b \in \X$ be two bona fide images of identities $a$, $b$,
and let $\bfx_0^{(a)} = x_a$ and $\bfx_0^{(b)} = x_b$.
\cref{alg:morphing} outlines the structure of the proposed Diffusion-based morphing attack, hereafter called the Diffusion attack for simplicity, with additional illustration provided in \cref{fig:diffae}.
We used 250 steps for the stochastic encoder and 100 steps for the DDIM sampler.

Beyond the core components of the DDIM and semantic encoder, three additional functions are added to the architecture,
namely, the image space preprocessing function, $\xi : \X \times \X \to \X$,
image space interpolation function, $\ell_\X : \X \times \X \times [0, 1] \to \X$,
and latent space interpolation function, $\ell_\Z : \Z \times \Z \times [0, 1] \to \Z$.

The interpolation functions are used to interpolate between the semantic and stochastic values by some factor $\gamma \in [0, 1]$.
The image space preprocessing function is used to prepare the image passed to the semantic encoder.
The simplest form of the interpolation function is the linear interpolation function, $\textrm{lerp}(a, b; \gamma) = \gamma a + (1 - \gamma) b$,
where linear interpolation was found to be the best choice for $\ell_\Z$.
However, Song~\etal~\cite{song2021denoising} suggests the usage of spherical linear interpolation for $\ell_\X$.
For a vector space $V$ and two vectors $u, v \in V$, the spherical interpolation by a factor of $\gamma$ is given as
\begin{equation}
    \textrm{slerp}(u, v; \gamma) = \frac{\sin((1 - \gamma) \theta)}{\sin \theta}u + \frac{\sin(\gamma \theta)}{\sin \theta}v
\end{equation}
where $\theta = \frac{\arccos(u \cdot v)}{\|u\| \, \|v\|}$.

While the semantic code provides most of the fundamental information, such as positioning of facial features, the stochastic code is used to provide information on the details not explicitly associated with the identity, but necessary for the realism of the generated image.
By altering the stochastic code, details such as direction of strands of hair, clothing, \textit{etc.}, are altered whilst the identity of the image is preserved \cite{diffae}.
Unlike the rather straightforward nature of linearly interpolating between the semantic codes to produce an image with key identifying characteristics of both identities, the nature of the stochastic code can lead to images of low visual fidelity if the interpolation is not done carefully.
In particular, linear interpolation between two stochastic codes does guarantee a smooth interpolation between the stochastic details in the images.
For this reason the preprocessing function, $\xi$, is used to prepare the image passed to the stochastic encoder.
One strategy is to ``pre-morph'' the image when extracting the stochastic details, \ie, $\xi$ performs an image space morph of the image with the goal of reducing the artifacts induced by stochastic interpolation.

\section{Experimental Setup}
To evaluate its effectiveness, we tested the proposed morphing attack on three datasets against two different state-of-the-art FR systems.
All training, optimization, and evaluation was conducted on a system with dual Intel Xeon Silver 4114 CPUs and an NVIDIA Tesla V100 32GB GPU with CUDA version 10.1 and CUDNN version 8.4.
The proposed morphing attack, MAD algorithm, and the FR systems are implemented in PyTorch~\cite{pytorch}.

\subsection{Face Recognition Systems}
In order to evaluate the strength of the proposed morphing attack, three publicly available FR systems are used, specifically, the FaceNet\footnote{\url{https://pypi.org/project/facenet-pytorch}}, VGGFace2\footnote{\url{https://github.com/ox-vgg/vgg_face2}}, and ArcFace\footnote{\url{https://github.com/deepinsight/insightface}} models.
These models are representative of recognition systems with state-of-the-art face verification performance~\cite{vggface2,facenet,deng2019arcface} and hence are widely used.
For both models the last fully connected layer is used to provide a rich feature representation of the input image.
Then for a presented face, its feature vector is compared with that of the feature vector belonging to the target face.
If the distance between these two representations is sufficiently ``small'', the presented face is then said to have the same identity as the target face.
The VGGFace2 model improves upon acclaimed VGGFace~\cite{vggface} by using an improved training dataset, also called VGGFace2.
Following the introduction of  Squeeze and Excitation Network (SENet) by Hu~\etal~\cite{squeezenet}, Cao~\etal~\cite{vggface2} presented the SENet architecture as the optimal choice when used with the VGGFace2 dataset.
Google's FaceNet model consists of an Inception-ResNet V1 architecture which is pre-trained on the VGGFace2 dataset~\cite{facenet}.
A new state-of-the-art model, ArcFace~\cite{deng2019arcface}, uses a novel loss function during training to improve the embeddings of faces.
This loss function is known as Additive Angular Margin Loss or ArcFace Loss.
The particular ArcFace model used for evaluation consists of a 100-layer Improved ResNet~\cite{iresnet} trained on the Glint360K dataset\footnote{The Glint360K dataset consists of 17,091,657 imagees of 360,232 individuals. Models trained on this dataset can achieve state-of-the-art performance.}~\cite{glint360k}.

Additionally, the three FR systems use different pre-processing pipelines.
Across all datasets the images and generated morphs are cropped as to be appropriate for passport photos; consequently, a face extractor such as MTCNN~\cite{mtcnn} is omitted from the verification pipeline.
The FaceNet model resizes the image such that the short side of the image is 180 pixels long and then the image is cropped to a $160 \times 160$ resolution.
Lastly, the images are normalized to $[-1, 1]$.
The VGGFace2 model resizes the image such that the short side of the image is 256 pixels long and then crops the image to $224 \times 224$ pixels.
The mean RGB vector\footnote{The mean vector is specifically $\langle 131.0912, 103.8827, 91.4953 \rangle$ for the red, green, and blue channels.}
is subtracted from the cropped image to normalize the image.
The ArcFace model resizes images such that the short side of the image is 112 pixels long which is then cropped to a $112 \times 112$ resolution. The image is then normalized to have values in $[-1, 1]$.

\subsection{Datasets}
\label{42dataset}
In this work, the FERET~\cite{feret}, FRLL~\cite{frll}, and FRGC v2.0~\cite{frgc} datasets were used to evaluate the proposed technique, as they are commonly used in MAD with a large number of different identities~\cite{sebastian_morphs, sebastian_gan_threaten}.
Notably, the FRLL dataset consists of high quality close-up frontal images at a $1350\times 1350$ resolution with 189 facial landmarks---a large number of landmarks.
The StyleGAN2, MIPGAN-II, and diffusion models were all trained on the FFHQ dataset, which contains 70,000 images at a $1024\times 1024$ resolution~\cite{stylegan}.
Morphs using OpenCV, FaceMorpher, and StyleGAN2 were created by Sarkar~\etal~\cite{sebastian_morphs} on the FRLL, FERET, and FRGC datasets.
Additionally, Zhang~\etal~\cite{mipgan} created morphs via MIPGAN-II on the three datasets.

In order to create a morphed face, two component identities are needed.
Naturally, if the two component identities are disparate, the resulting morph is likely to be very weak.
To rectify this and for evaluation purposes, the component identity pairs were selected by following the existing protocol offered by Sarkar~\etal~\cite{sebastian_morphs}.
These pairings resulted in 1222 unique morphs on FRLL, 964 on FRGC, and 529 on FERET.

\section{Results}

\begin{figure*}[ht!]
    \centering
    \begin{subfigure}[t]{0.13\textwidth}
        \includegraphics[width=\textwidth, trim={1.3cm 0 1.3cm 0.2cm}, clip]{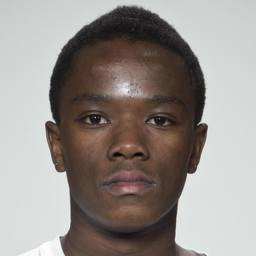}
        \caption{Identity $a$}
        \label{fig:morphs-a}
    \end{subfigure}
    \begin{subfigure}[t]{0.13\textwidth}
        \includegraphics[width=\textwidth, trim={1.2cm 0.4cm 1.2cm 0cm}, clip]{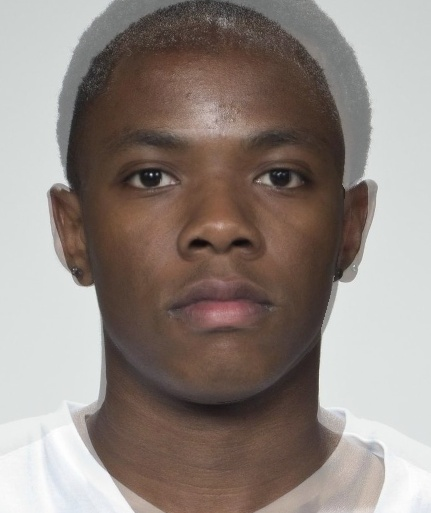}
        \caption{OpenCV}
        \label{fig:morphs-opencv}
    \end{subfigure}
    \begin{subfigure}[t]{0.13\textwidth}
        \includegraphics[width=\textwidth, trim={6cm 0 6cm 2.75cm}, clip]{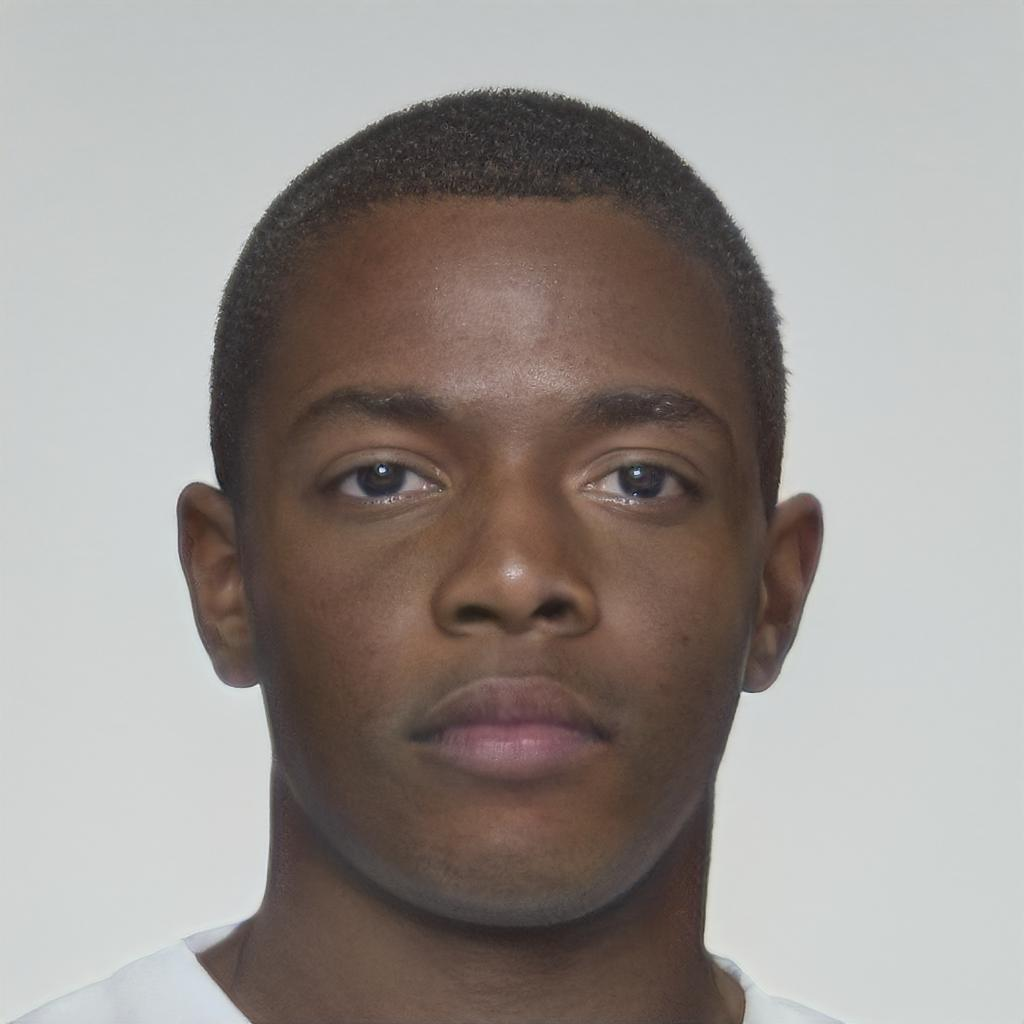}
        \caption{StyleGAN2}
        \label{fig:morphs-stylegan}
    \end{subfigure}
    \begin{subfigure}[t]{0.13\textwidth}
        \includegraphics[width=\textwidth, trim={1.6cm 0 1.6cm 1cm}, clip]{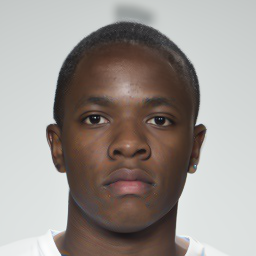}
        \caption{Diffusion}
        \label{fig:morphs-diffusion}
    \end{subfigure}
    \begin{subfigure}[t]{0.13\textwidth}
        \includegraphics[width=\textwidth, trim={6cm 0 6cm 2.75cm}, clip]{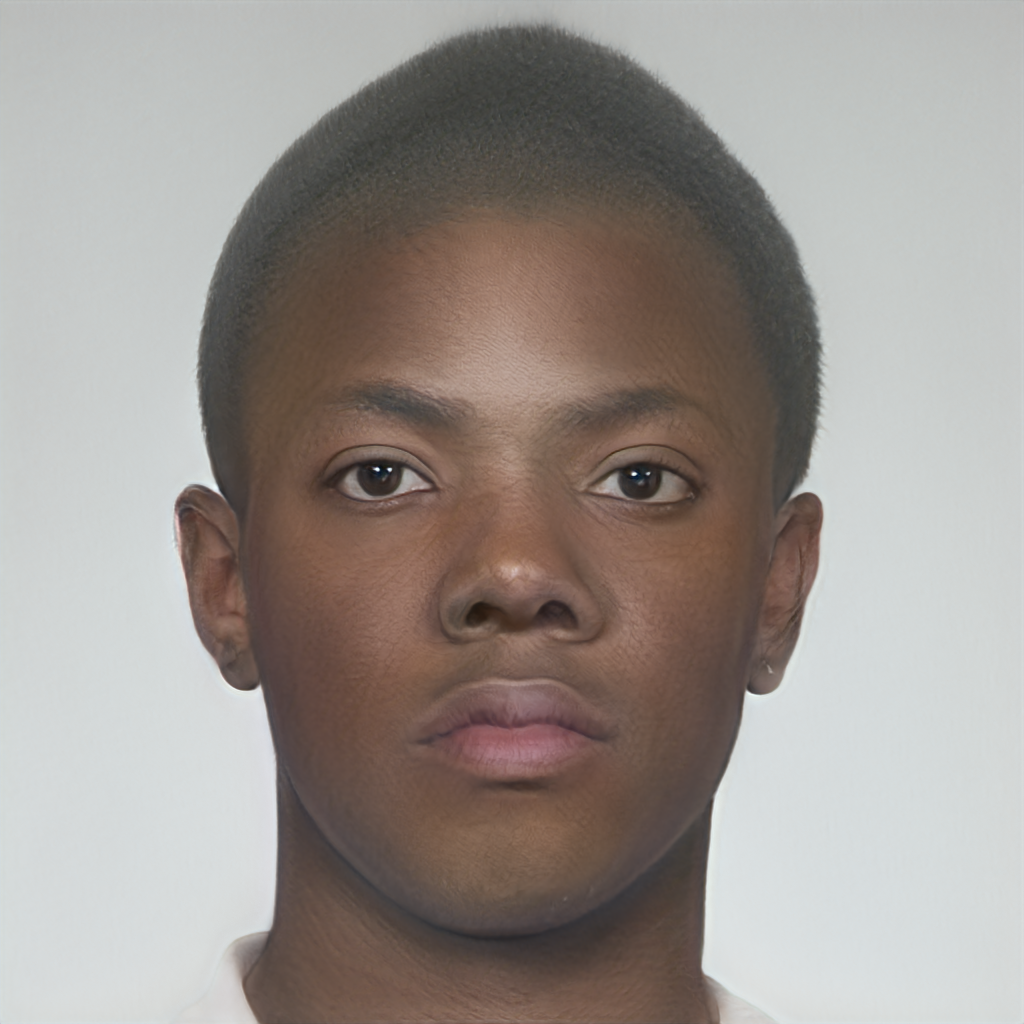}
        \caption{MIPGAN-II}
        \label{fig:morphs-mipgan}
    \end{subfigure}
    \begin{subfigure}[t]{0.13\textwidth}
        \includegraphics[width=\textwidth, trim={0.8cm 0cm 0.8cm 0cm}, clip]{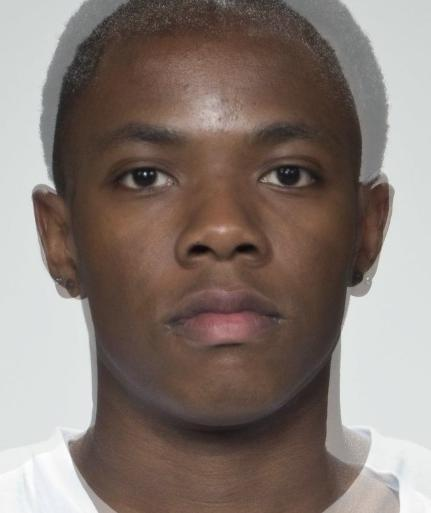}
        \caption{FaceMorpher}
        \label{fig:morphs-facemorpher}
    \end{subfigure}
    \begin{subfigure}[t]{0.13\textwidth}
        \includegraphics[width=\textwidth, trim={1.4cm 0 1.4cm 0.3cm}, clip]{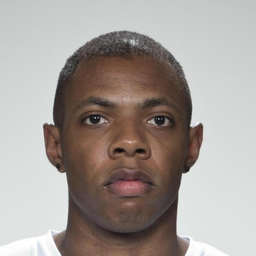}
        \caption{Identity $b$}
        \label{fig:morphs-b}
    \end{subfigure}
    
    \begin{subfigure}[t]{0.13\textwidth}
        \includegraphics[width=\textwidth, trim={1.14cm 0 1.14cm 0cm}, clip]{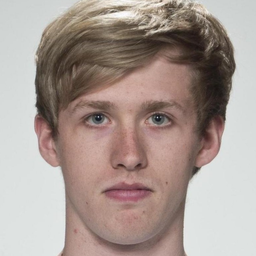}
        \caption{Identity $c$}
        \label{fig:2morphs-c}
    \end{subfigure}
    \begin{subfigure}[t]{0.13\textwidth}
        \includegraphics[width=\textwidth, trim={1.22cm 1cm 1.22cm 0cm}, clip]{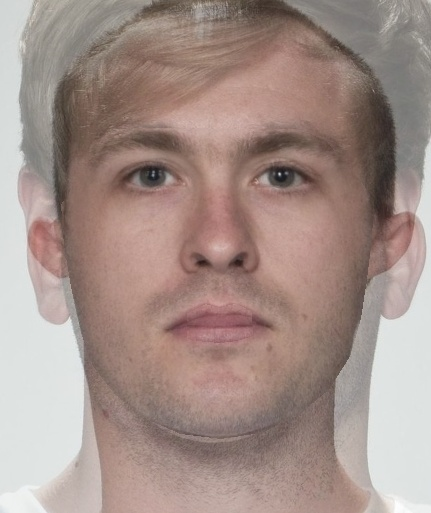}
        \caption{OpenCV}
        \label{fig:2morphs-opencv}
    \end{subfigure}
    \begin{subfigure}[t]{0.13\textwidth}
        \includegraphics[width=\textwidth, trim={5cm 0 5cm 1cm}, clip]{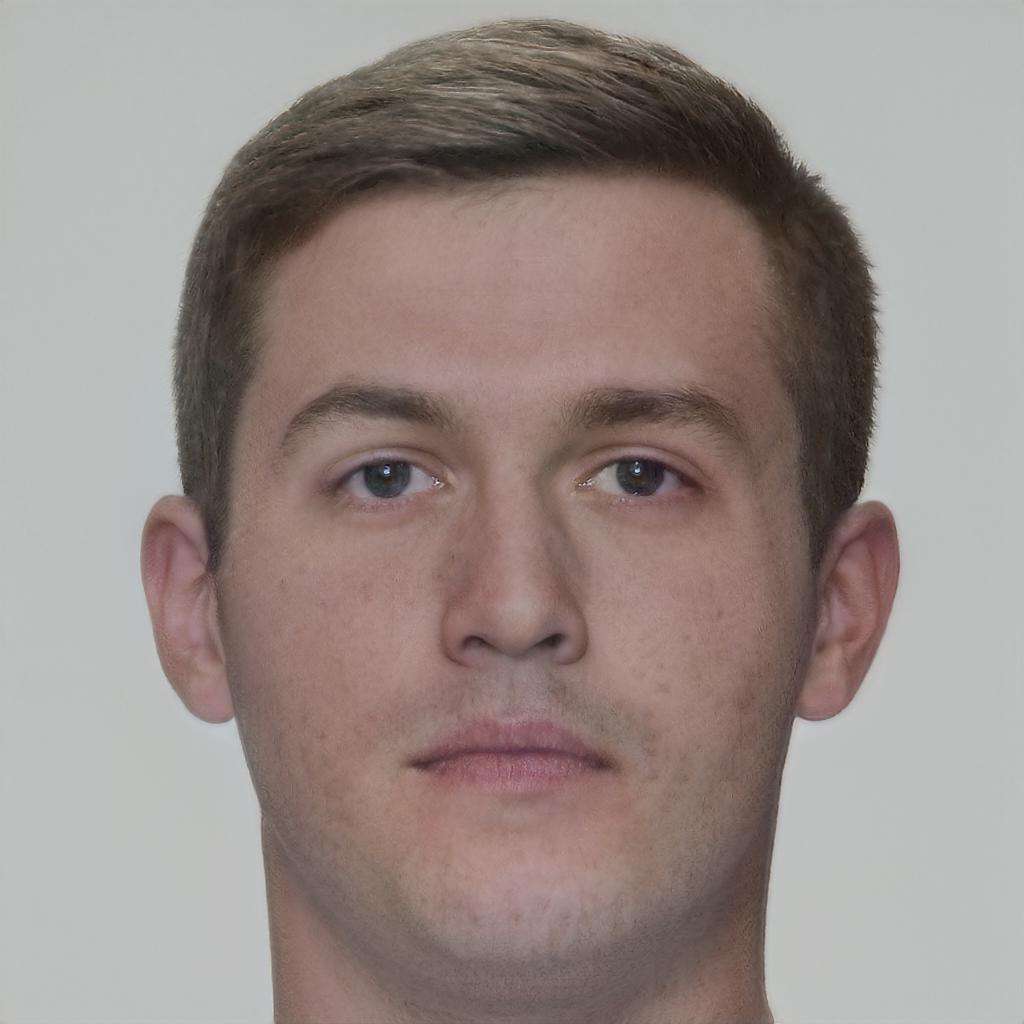}
        \caption{StyleGAN2}
        \label{fig:2morphs-stylegan}
    \end{subfigure}
    \begin{subfigure}[t]{0.13\textwidth}
        \includegraphics[width=\textwidth, trim={1.15cm 0 1.15cm 0}, clip]{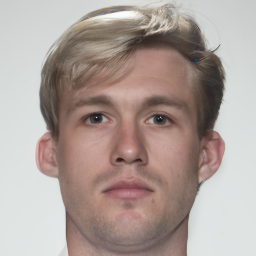}
        \caption{Diffusion}
        \label{fig:2morphs-diffusion}
    \end{subfigure}
    \begin{subfigure}[t]{0.13\textwidth}
        \includegraphics[width=\textwidth, trim={4.7cm 0 4.7cm 0}, clip]{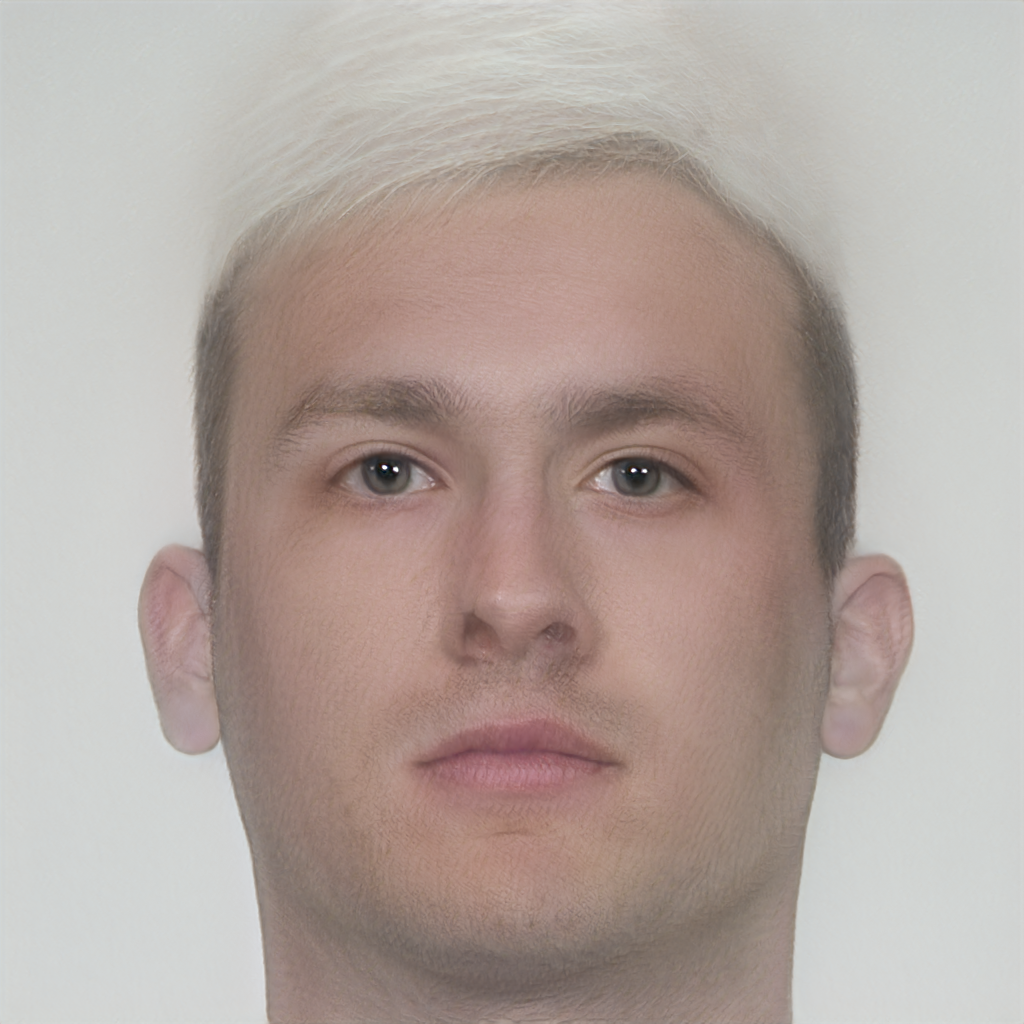}
        \caption{MIPGAN-II}
        \label{fig:2morphs-mipgan}
    \end{subfigure}
    \begin{subfigure}[t]{0.13\textwidth}
        \includegraphics[width=\textwidth, trim={1.02cm 1cm 1.02cm 0cm}, clip]{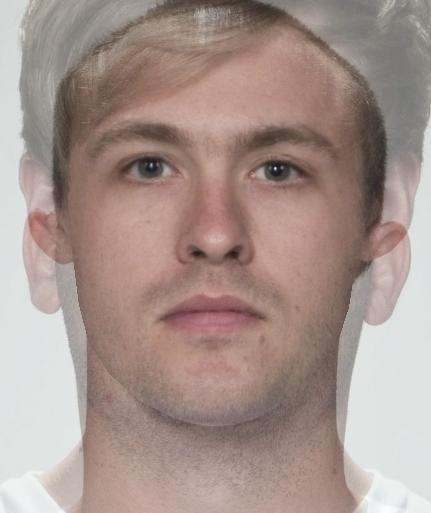}
        \caption{FaceMorpher}
        \label{fig:2morphs-facemorpher}
    \end{subfigure}
    \begin{subfigure}[t]{0.13\textwidth}
        \includegraphics[width=\textwidth, trim={1.17cm 0 1.17cm 0}, clip]{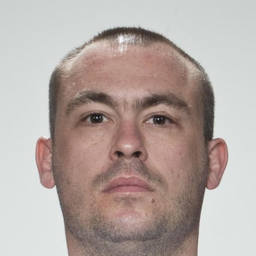}
        \caption{Identity $d$}
        \label{fig:2morphs-d}
    \end{subfigure}
    \caption{Comparison across different morphing algorithms of two identity pairs from the FRLL dataset.} 
    \label{fig:morphs}
\end{figure*}

\begin{figure}
    \centering
    \begin{subfigure}[t]{0.24\textwidth}
        \includegraphics[width=\textwidth, trim={32px 32px 32px 32px}, clip]{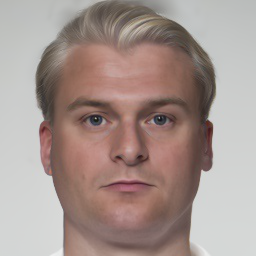}
        \caption{Diffusion}
        \label{fig:diff_mipgan_diff}
    \end{subfigure}
    \begin{subfigure}[t]{0.24\textwidth}
        \includegraphics[width=\textwidth, trim={32px 32px 32px 32px}, clip]{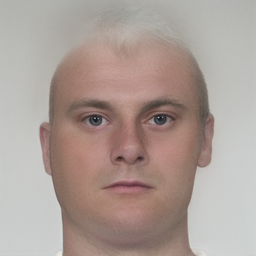}
        \caption{MIPGAN-II}
        \label{fig:diff_mipgan_mipgan}
    \end{subfigure}
    \caption{Comparison of Diffusion and MIPGAN-II morphed faces on FRLL. Images are resized to $256\times256$ and cropped to $224\times224$ to match VGGFace2 pre-processing pipeline.}
    \label{fig:diff_vs_mipgan}
\end{figure}

The proposed morphing attack is compared to state-of-the-art techniques drawing from both GAN-based and Landmark-based methods.
The effectiveness of the proposed method is quantitatively measured on three fronts, being the visual fidelity of the generated morphed images, the vulnerability of state-of-the-art FR systems to the morphing attack, and the detection potential of the morphing attack, respectively.
Furthermore, an exploration of interpolation techniques for the stochastic latent code is provided.

\begin{table}[t]
    \centering
    \caption{FID across different morphing attacks. Lower is better.}
    \begin{tabular}{lrrr}
    \toprule
    \textbf{Morphing Attack}  &  \textbf{FRLL} &   \textbf{FRGC} &  \textbf{FERET} \\
    \midrule
    StyleGAN2 & 45.19 & 86.41 & \textbf{41.91}\\
    FaceMorpher & 91.97 & 88.14 & 79.58\\
    OpenCV & 85.71 & 100.02 & 91.94\\
    MIPGAN-II & 66.41 & 115.96 & 70.88\\
    Diffusion & \textbf{42.63} & \textbf{64.16} & 50.45\\
    \bottomrule
    \end{tabular}
    \label{tab:fid_all}
\end{table}

\begin{table*}[t]
\centering
\caption{The APCER at specific BPCER values. Higher is better.}
\begin{tabular}{lllrrr}
\toprule
\textbf{Dataset}   & \textbf{FR System}   & \textbf{Morphing Attack}       &   \textbf{APCER @ BPCER = 0.1\%} & \textbf{APCER @ BPCER = 1\%} & \textbf{APCER @ BPCER = 5\%} \\
\midrule
 \multirow{15}{*}{FRLL}      & FaceNet     & StyleGAN2   &                 0.99 &                 0.05 &                 0    \\
       & FaceNet     & FaceMorpher &                 2.25 &                 0.14 &                 0.05 \\
       & FaceNet     & OpenCV      &                 3.24 &                 0.33 &                 0    \\
       & FaceNet     & MIPGAN-II   &                 \textbf{8.87} &                 0.47 &                 0.09 \\
       & FaceNet     & Diffusion   &                 8.83 &                 \textbf{0.99} &                 \textbf{0.23} \\
       \addlinespace
       & VGGFace2    & StyleGAN2   &                 0.05 &                 0.05 &                 0    \\
       & VGGFace2    & FaceMorpher &                 1.36 &                 1.08 &                 0.23 \\
       & VGGFace2    & OpenCV      &                 2.35 &                 \textbf{2.11} &                 0.28 \\
       & VGGFace2    & MIPGAN-II   &                 1.31 &                 0.99 &                 0.23 \\
       & VGGFace2    & Diffusion   &                 \textbf{2.68} &                 2.07 &                 \textbf{0.52} \\
       \addlinespace
       & ArcFace     & StyleGAN2   &                 0    &                 0    &                 0    \\
       & ArcFace     & FaceMorpher &                 0    &                 0    &                 0    \\
       & ArcFace     & OpenCV      &                 0    &                 0    &                 0    \\
       & ArcFace     & MIPGAN-II   &                 0.05 &                 0    &                 0    \\
       & ArcFace     & Diffusion   &                 \textbf{3.33} &                 \textbf{0.28} &                 \textbf{0.09} \\
 \midrule
 \multirow{15}{*}{FRGC}      & FaceNet     & StyleGAN2   &                74.04 &                36.69 &                17.73 \\
       & FaceNet     & FaceMorpher &                87.9  &                38.85 &                14.9  \\
       & FaceNet     & OpenCV      &                84.1  &                31.43 &                11.36 \\
       & FaceNet     & MIPGAN-II   &                \textbf{96.54} &                \textbf{61.9}  &                \textbf{33.48} \\
       & FaceNet     & Diffusion   &                91.73 &                48.86 &                24.95 \\
       \addlinespace
       & VGGFace2    & StyleGAN2   &                81.42 &                46.22 &                26.7  \\
       & VGGFace2    & FaceMorpher &                95.42 &                63.65 &                38.18 \\
       & VGGFace2    & OpenCV      &                \textbf{95.31} &                \textbf{64.62} &                \textbf{39.4}  \\
       & VGGFace2    & MIPGAN-II   &                91.92 &                57.8  &                30.84 \\
       & VGGFace2    & Diffusion   &                93.71 &                58.25 &                32.18 \\
       \addlinespace
       & ArcFace     & StyleGAN2   &                18.44 &                 6.26 &                 0.93 \\
       & ArcFace     & FaceMorpher &                26.11 &                 5.44 &                 0.37 \\
       & ArcFace     & OpenCV      &                 2.12 &                 0.07 &                 0    \\
       & ArcFace     & MIPGAN-II   &                30.39 &                12.18 &                 2.79 \\
       & ArcFace     & Diffusion   &                \textbf{40.6}  &                \textbf{20.41} &                 \textbf{7.49} \\
\midrule
 \multirow{15}{*}{FERET}     & FaceNet     & StyleGAN2   &                15.65 &                 9.35 &                 2.49 \\
      & FaceNet     & FaceMorpher &                10.71 &                 5.1  &                 0.91 \\
      & FaceNet     & OpenCV      &                 8.79 &                 3.06 &                 0.17 \\
      & FaceNet     & MIPGAN-II   &                21.03 &                10.77 &                 2.21 \\
      & FaceNet     & Diffusion   &                \textbf{24.04} &                \textbf{13.95} &                 \textbf{4.99} \\
       \addlinespace
      & VGGFace2    & StyleGAN2   &                54.08 &                18.42 &                 5.73 \\
      & VGGFace2    & FaceMorpher &                80.5  &                32.65 &                12.7  \\
      & VGGFace2    & OpenCV      &                \textbf{81.01} &                32.6  &                12.87 \\
      & VGGFace2    & MIPGAN-II   &                66.5  &                18.14 &                 5.84 \\
      & VGGFace2    & Diffusion   &                80.9  &                \textbf{35.2}  &                \textbf{14.34} \\
       \addlinespace
      & ArcFace     & StyleGAN2   &                 0.51 &                 0    &                 0    \\
      & ArcFace     & FaceMorpher &                 0.17 &                 0    &                 0    \\
      & ArcFace     & OpenCV      &                 0    &                 0    &                 0    \\
      & ArcFace     & MIPGAN-II   &                 0.06 &                 0    &                 0    \\
      & ArcFace     & Diffusion   &                 \textbf{9.69} &                 \textbf{2.27} &                 \textbf{0.06} \\
\bottomrule
\end{tabular}
\label{tab:det}
\end{table*}

\subsection{Evaluation of Visual Fidelity}
The visual fidelity of the Diffusion attack is compared against other morphing attacks.
Whilst on first glance it may appear that the ability to deceive an FR system should imply a high level of visual fidelity, this is not a simple assertion.
We posit two reasons for this discrepancy:
\begin{enumerate}
    \item The image pre-processing pipeline for an FR system may crop out a significant portion of artifacts in the original morphed image.
    \item The FR system can be vulnerable to certain adversarial morphing attacks even though they inject noticeable artifacts into the morphed image.
\end{enumerate}
This can lead to a situation wherein the FR system is fooled by a morphed image; however, it would be trivial for a human agent to notice the artifacts present in the image.
Moreover, a deep learning system could be specifically trained to notice such artifacts, greatly reducing the potential of such an attack to go undetected.

To quantitatively assess the visual fidelity of the generated images, the
Fr\'echet Inception Distance (FID) is employed, as it has shown to correlate well with human assessment of fidelity~\cite{FID}.
The FID is a measure of distance between the generated and target distributions. Therefore, the lower the FID metric, the more similar the generated distribution is to the target distribution, which correlates well with visual fidelity.
The metric is defined as the Fr\'echet distance, or 2-Wasserstein metric\footnote{The 2-Wasserstein metric between two probability measures $\mu, \nu$ with finite moments on $\R^n$ is defined as
\begin{equation*}
    W_2(\mu, \nu) = \bigg(\inf_{\pi \in \Pi(\mu, \nu)} \int_{\R^n \times \R^n} \|x-y\|_2^2\;\textrm{d}\pi(x, y)\bigg)^\frac 12
\end{equation*}
where $\Pi(\mu, \nu)$ is the set of all distributions with marginals $\mu$ and $\nu$.
},
between two Gaussian distributions, each representing the activations the deepest layer of an Inception v3 network\footnote{The Inception v3 network is trained on the publically available ImageNet dataset~\cite{imagenet}.} induced by images from the generated and target distributions.

\cref{tab:fid_all} shows the FID metric between the generated images from different morphing attacks and bona fide samples from the dataset the morphing attack is drawn from.
The FID metric is calculated using \texttt{pytorch-fid}~\cite{Seitzer2020FID}.
What we can see is that morphed images generated from the Diffusion attack generally has the lowest FID, with the StyleGAN2-based attack following closely behind.
Both Landmark-based morphs---OpenCV and FaceMorpher---has noticeably higher FIDs than the deep learning-based morphs.
These results correlate well with visual inspection of the morphs as \cref{fig:morphs-opencv,fig:morphs-facemorpher,fig:2morphs-opencv,fig:2morphs-facemorpher} exhibit prominent artifacts outside the central face region.
Likewise, the MIPGAN-II attack seems to struggle with some distortion outside the central face region, as can be seen in \cref{fig:morphs-mipgan,fig:2morphs-mipgan}.
Interestingly, on the FRLL dataset the StyleGAN2 morphing pipeline consistently darkens morphs relative to its component images; however, the visual fidelity is relatively high albeit noticeable darkening.
Importantly, stochastic details such as hair seem to be modelled well by the Diffusion attack, while other attacks distort such details for the cases of OpenCV, FaceMorpher, and MIPGAN-II attacks, or present details that have little similarity to both identities for the case of StyleGAN2 attack.
While exhibiting far less visual artifacts than other morphing techniques, the Diffusion attack tends to slightly smooth out the skin texture.
Overall, the Diffusion attack exhibits the highest consistent visual fidelity among all presented attacks.

Interestingly, the MIPGAN-II morphs exhibit a much higher FID than the StyleGAN2 morphs,
despite the two approach being based on the same deep learning backbone, \ie, the StyleGAN2 model.
\cref{fig:diff_vs_mipgan} compares two morphed faces generated by the Diffusion and MIPGAN-II attacks on the FRLL dataset.
High frequency artifacts can be easily observed in \cref{fig:diff_mipgan_mipgan}, particularly, near the hairline and the transition region between hair and the background.
Comparing \cref{fig:diff_mipgan_diff} and \cref{fig:diff_mipgan_mipgan}, the hair generated by the MIPGAN-II attack looks unnatural with a strange texture as though an image sharpening filter has been applied to the image, greatly enhancing the magnitude of high frequency content,
which aligns with the observation in \cref{fig:morphs}.
Moreover, the MIPGAN-II images seem to be de-saturated when compared to images produced by other attacks, leading to a washed-out appearance.
One possible explanation for the low visual fidelity is the identity loss overpowering the perceptual quality loss, leading to morphed images with low visual fidelity but high effectiveness against FR systems.

\subsection{Vulnerability of FR Systems}
The strength of the proposed face morphing algorithm is further evaluated by measuring the ability of the morph to deceive an FR system.
The attack success is quantitatively verified against two state-of-the-art FR systems.
To ensure a valid comparison across five different morphing attacks, the same pairs of component identities were used in evaluating every morphing attack, \ie, for every pair of component identities a morphed image was created for each of the five attacks.
For the FaceNet, VGGFace2, and ArcFace FR systems, the False Match Rate (FMR) is set at 0.1\% following the guidelines of Frontex~\cite{frontex}.
Additionally, the distance between faces is measured using the $L^2$ distance between the outputs of the FR model.
All measurements were collected by embedding the bona fide, imposter, and morphed images from all morphing attacks using the three FR systems.
Any derivative metrics of performance were calculated using the embeddings we collected.

The vulnerability of FR systems to morphing attacks is assessed by comparing the error rates in detection. Specifically, the Attack Presentation Classification Error Rate (APCER)\footnote{APCER is the proportion of attack presentations incorrectly classified as bona fide presentations.}
is measured at specific Bona fide Presentation Classification Error Rate (BPCER)\footnote{BPCER is the proportion of bona fide presentations incorrectly classified as attack presentations.}
values.
The FR systems are treated as differential morph detectors, \ie, two images are sent to the FR model, one of which is bona fide and the other is unknown.
The $L^2$ distance between the embeddings of bona fide image and the unknown image is calculated.
A threshold is set to obtain various desired BPCER values. If the distance is less than this threshold the unknown sample is classified as bona fide; otherwise, it is not.
In \cref{tab:det} the APCER values for the five different morphing attacks are presented across all three datasets evaluated on three different BPCER values of 0.1\%, 1\%, and 5\%, respectively.
\begin{table*}[t]
    \centering
    \caption{MMPMR at FMR = 0.1\% across different morphing attacks. Higher is better. \textdagger~the geometric mean.}
    \begin{tabular}{lrrrrrrrrrr}
        \toprule
        & \multicolumn{3}{c}{\textbf{FRLL}}
        & \multicolumn{3}{c}{\textbf{FRGC}}
        & \multicolumn{3}{c}{\textbf{FERET}}\\
        \cmidrule(lr){2-4}
        \cmidrule(lr){5-7}
        \cmidrule(lr){8-10}
        \textbf{Morphing Attack} &
        \textbf{FaceNet} & \textbf{VGGFace2} & \textbf{ArcFace} &
        \textbf{FaceNet} & \textbf{VGGFace2} & \textbf{ArcFace} &
        \textbf{FaceNet} & \textbf{VGGFace2} & \textbf{ArcFace} &
        \textbf{Mean\textsuperscript{\textdagger}}\\
        \midrule
         StyleGAN2         &           4.69 &         6.05 &          19.89 &           0.18 &         0.85 &           5.49 &            0.54 &          0.76 &            4.95 & 2.15\\
         FaceMorpher       &          11.26 &        36.4  &          45.03 &           0.51 &         9.15 &          41.28 &            2.3  &         10.78 &           \textbf{60.73} & 12.05\\
         OpenCV            &          17.34 &        \textbf{40.93} &          47.7  &           0.14 &        \textbf{12.16} &           3.99 &            1.69 &         11.12 &            4.61 & 6.47\\
         MIPGAN-II         &          \textbf{30.96} &        26.74 &          56.52 &           \textbf{3.12} &         7.94 &          33.54 &            6    &          5.39 &           18.19 & 14.16\\
         Diffusion         &          28.14 &        35.37 &          \textbf{88.09} &           2.68 &         8.47 &          \textbf{46.74} &            \textbf{6.47} &         \textbf{13.03} &           59.75 & \textbf{19.80}\\
        \bottomrule
    \end{tabular}
    \label{tab:mmpmr}
\end{table*}
Due to a variety of factors---such as image quality and number of bona fide images per identity---the results vary across different datasets; while there is some variance across different FR systems, they tend to agree more closely.
Noticeably, all attacks perform rather poorly on the FRLL dataset, although the Diffusion attack performs the best among them, which could be attributed to the limited number of bona fide images per identity;
for FRLL dataset there are only two bona fide images per identity: a neutral face (used to create the morph) and a smiling face.
All FR systems were more vulnerable to the different morphing attacks when evaluated on the FRGC dataset.
The MIPGAN-II attack performed very well against FaceNet on the FRGC dataset, which makes sense as this technique was refined on the FRGC dataset in particular~\cite{mipgan}.
However, this attack was not as strong against VGGFace2. Instead, that FR system was more vulnerable to OpenCV, FaceMorpher, and Diffusion. This could be attributed to the different pre-processing pipelines.
The Diffusion-based morphing attack generally performs close to the top performer on all FR systems.
As with FRGC, on the FERET dataset VGGFace2 is more vulnerable to landmark-based attacks, OpenCV and FaceMorpher, than FaceNet.
Diffusion-based morphs pose the greatest threat on FERET, consistently having high APCER values.
In general, the following observations can be drawn from \cref{tab:det}:
\begin{itemize}
    \item Among the five different attacks, FR systems are most vulnerable to Diffusion attacks. Moreover, Diffusion attacks always rank in the top three in terms of performance.
    \item FR systems are the least vulnerable to the StyleGAN2 attack. The StyleGAN2 attack is consistently outperformed by its successor, MIPGAN-II, and the other deep learning-based attack, Diffusion, while often falling behind landmark-based attacks.
    \item Even though the state-of-the-art ArcFace model is very resilient to morphing attacks, it presents a significant vulnerability to Diffusion morphs, compared to other morphing attacks.
\end{itemize}
In addition to using the error rates to assess the vulnerability of FR systems, the Mated Morphed Presentation Match Rate (MMPMR)~\cite{mmpmr} is used as a measure of vulnerability.
Scherhag~\etal~\cite{mmpmr} proposed two variants of the MMPMR metric for the scenario in which multiple bona fide images of an identity were used in morph process, excluding the image used to create the morph, called the MinMax-MMPMR and ProdAvg-MMPMR.
The MinMax-MMPMR metric is likely to increase the number of accepted morphs as the number of bona fide images per identity increases.
Therefore, the ProdAvg-MMPMR is the specific MMPMR variant used to assess the vulnerability of FR systems. Any mention hereafter to MMPMR refers specifically to ProdAvg-MMPMR unless stated otherwise.

Let $\pr_M \in \mathcal{P}(\X)$ be the distribution of morphed images such that for some $x_{ab} \sim \pr_M$, $x_{ab}$ denotes a morphed image made from identities $a, b$, where $\mathcal{P}(\X)$ denotes the set of all probability measures on $\X$.
Let $\pr_k \in \mathcal{P}(\X)$ denote the distribution of bona fide images of identity $k$.
Then with abuse of notation $\pr_{k \setminus x_{ab}}$ is the distribution of bona fide images of identity $k$ excluding those images used in creating the morph $x_{ab}$.
The MMPMR metric for a particular threshold, $\gamma > 0$, equipped with FR system $F: \X \to V$ is then defined as
\begin{equation*}
    M(\gamma) = \ex_{x_{ab} \sim \pr_M}\Bigg[ \prod_{k \in \{a, b\}} \ex_{x \sim \pr_{k\setminus x_{ab}}}\Big[\|F(x_{ab}) - F(x)\|_2 < \gamma\Big] \Bigg]
\end{equation*}
\ie, the expected success rate of the morphing attack to deceive the FR system.
The product term is the joint probability of successful verification of both identities.

\cref{tab:mmpmr} presents the MMPMR metric when the FMR is set at 0.1\% for all datasets and FR systems.
Interestingly, the FRLL dataset has the highest overall MMPMR metrics in contrast to the results from~\cref{tab:det}.
This can likely be attributed to limited number of bona fide images per identity in the FRLL dataset, in contrast with other datasets, as the particular choice of MMPMR metric heavily punishes failed verifications for either identity.
Therefore, with FRLL only having one possible bona fide image per identity the MMPMR metric could be skewed higher relative to the other datasets.
On average the Diffusion attack greatly outperforms the other attacks; conversely, the Landmark-based attacks on average exhibit mediocre performance.
In agreement with \cref{tab:det} the StyleGAN2 attack shows abysmal performance in comparison with the other attacks.
It is noteworthy the state-of-the-art ArcFace model seems more vulnerable than the other FR systems, and is particular vulnerable to the Diffusion attack.

\begin{table}[]
    \centering
    \caption{APCER at FMR = 0.1\% across different margin sizes on the FaceNet FR system. Higher is Better.}
    \begin{tabular}{llrrrr}
    \toprule
    & & \multicolumn{4}{c}{\textbf{Margin Size}}\\
    \cmidrule(lr){3-6}
    \textbf{Dataset}   & \textbf{Morphing Attack}   &     \textbf{0} &    \textbf{20} &    \textbf{40} &    \textbf{80} \\
    \midrule
    \multirow{5}{*}{FRLL}      & MIPGAN-II         & 54.84 & 56.53 & 57.18 & 58.03 \\
          & StyleGAN2         & 15.12 & 15.57 & 17.14 & 25.11 \\
          & FaceMorpher       & 74.48 & 75.26 & 73.86 & 47.91 \\
          & OpenCV            & 76.21 & 75.82 & 74.96 & 48.4  \\
          & Diffusion         & 51.25 & 54.47 & 57.07 & 59.02 \\
    \addlinespace
    \multirow{5}{*}{FERET}     & MIPGAN-II         & 19.33 & 21.71 & 22.11 & 26.36 \\
         & StyleGAN2         & 14.12 & 17.57 & 18.14 & 22.17 \\
         & FaceMorpher       & 36.11 & 36.45 & 36.28 & 17.8  \\
         & OpenCV            & 36.11 & 38.44 & 37.64 & 13.89 \\
         & Diffusion         & 23.24 & 26.02 & 25.91 & 30.39 \\
    \addlinespace
    \multirow{5}{*}{FRGC}      & MIPGAN-II         & 12.33 & 14.3  & 16.2  & 20.97 \\
          & StyleGAN2         &  7.18 &  8.71 &  9.74 & 14.42 \\
          & FaceMorpher       & 17.5  & 18.87 & 20.39 &  9.07 \\
          & OpenCV            & 17.02 & 18.47 & 19.91 &  6.6  \\
          & Diffusion         &  9.7  & 10.71 & 12.27 & 15.62 \\
    \bottomrule
    \end{tabular}
    \label{tab:margin_size}
\end{table}

\subsubsection{The Effect of Pre-processing on an FR System}
\label{sec:margins}
Here the impact of the pre-processing pipeline on the vulnerability of an FR system is examined.
In particular the cropping process is further explored.
To study this an additional margin size is added to the image after an initial face extraction and cropping performed by MTCNN, such that a margin size $N$ adds back at most $N$ pixels to the cropped image in both dimensions.
Therefore, the larger $N$ is the less tightly cropped the image passed to the FR system is.
\cref{tab:margin_size} illustrates the impact of the margin size on the APCER metric on the FaceNet FR system.
Generally, as the margin size increases, the performance of Landmark-based attacks decreases while the performance of deep learning-based attacks increases.
As illustrated in \cref{fig:morphs}, the Landmark-based attacks have noticeable artifacts outside the central face region;
conversely, the deep learning-based morphs have less artifacts in the outside regions and generally look more realistic to a human observer.
This observation aligns with the visual fidelity results from \cref{tab:fid_all}.
Therefore, a MAD algorithm or FR system which uses less tightly cropped faces would be more resilient against attacks with visual artifacts outside the core face region.

\begin{table*}[t]
    \centering
    \caption{Ablation study on the impact morphing attack on validation accuracy.}
    \begin{adjustbox}{width=\linewidth}
    \begin{tabular}{lcccccrrrrr}
    \toprule
    & \multicolumn{5}{c}{\textbf{Training Attack}} & \multicolumn{5}{c}{\textbf{Validation Attack}}\\
    \cmidrule(lr){2-6}
    \cmidrule(lr){7-11}
     \textbf{Dataset}   & \textbf{Diffusion}   & \textbf{FaceMorpher}   & \textbf{MIPGAN-II}   & \textbf{OpenCV}    & \textbf{StyleGAN2}   &   \textbf{Diffusion} &   \textbf{FaceMorpher} &   \textbf{MIPGAN-II} &   \textbf{OpenCV} &   \textbf{StyleGAN2} \\
    \midrule
     \multirow{5}{*}{FERET}     & \xmark       & \cmark     & \cmark   & \cmark & \cmark   &       72.73 &         99.23 &      100    &    99.95 &       99.33 \\
          & \cmark   & \xmark         & \cmark   & \cmark & \cmark   &       99.9  &         76.39 &      100    &    99.85 &       99.64 \\
          & \cmark   & \cmark     & \xmark       & \cmark & \cmark   &       99.69 &         99.38 &      100    &    99.95 &       99.54 \\
          & \cmark   & \cmark     & \cmark   & \xmark     & \cmark   &       99.74 &         99.48 &      100    &    99.74 &       99.43 \\
          & \cmark   & \cmark     & \cmark   & \cmark & \xmark       &       99.74 &         98.56 &       99.9  &    99.74 &       87.89 \\
     \addlinespace
     \multirow{5}{*}{FRGC}      & \xmark       & \cmark     & \cmark   & \cmark & \cmark   &       75.89 &         99.98 &       99.97 &    99.9  &       99.93 \\
           & \cmark   & \xmark         & \cmark   & \cmark & \cmark   &       99.95 &         99.48 &      100    &    99.9  &       99.95 \\
           & \cmark   & \cmark     & \xmark       & \cmark & \cmark   &       99.83 &         99.85 &       99.82 &    99.8  &       99.85 \\
           & \cmark   & \cmark     & \cmark   & \xmark     & \cmark   &       99.93 &        100    &      100    &    99.23 &       99.93 \\
           & \cmark   & \cmark     & \cmark   & \cmark & \xmark       &       99.93 &         99.93 &       99.94 &    99.88 &       97.83 \\
     \addlinespace
     \multirow{5}{*}{FRLL}      & \xmark       & \cmark     & \cmark   & \cmark & \cmark   &       13.96 &         99.58 &       99.32 &    99.65 &       99.65 \\
           & \cmark   & \xmark         & \cmark   & \cmark & \cmark   &       99.23 &         99.09 &       98.91 &    99.37 &       99.44 \\
           & \cmark   & \cmark     & \xmark       & \cmark & \cmark   &       99.09 &         98.95 &       98.24 &    99.02 &       99.09 \\
           & \cmark   & \cmark     & \cmark   & \xmark     & \cmark   &       99.51 &         99.44 &       99.19 &    99.16 &       99.58 \\
           & \cmark   & \cmark     & \cmark   & \cmark & \xmark       &       99.93 &         99.86 &       99.86 &    99.93 &       95.02 \\
    \bottomrule
    \end{tabular}
    \end{adjustbox}
    \label{tab:holdout}
\end{table*}

\subsubsection{General Remarks on the Vulnerability Study}
The poor performance of the StyleGAN2 attack could be attributed to the darkening of images with light backgrounds, see \cref{fig:morphs}, and the aliasing effects latent to the StyleGAN2 generation pipeline, which was addressed by Karras~\etal~\cite{stylegan3}.
Moreover, the structure of the StyleGAN2 latent space can make exploration in the space difficult, which could possibly explain the poor performance in attacking the FR system compared to other attacks.
MIPGAN-II, on the other hand, likely avoids these pitfalls due to its explicit latent optimization process for deceiving an FR system.
The Diffusion attack utilizes an entirely different latent representation scheme, which seems to yield an advantage in the task of generating morphed faces.
The pre-processing pipeline of the FR system seems to mostly mitigate the artifacts latent to the Landmark-based attacks;
however, such artifacts could easily be detected by a human observer.
Lastly, the state-of-the-art ArcFace FR system seems especially vulnerable to the Diffusion attack.

\subsection{Detectability of Morphing Attacks}
The performance of the proposed attack is further evaluated by the ability of Morphing Attack Detection (MAD) algorithms trained against other attacks to detect an unseen attack.
To quantitatively assess the detectability of a particular morphing attack, a
SE-ResNeXt101-32x4d model pre-trained on ImageNet~\cite{imagenet} by NVIDIA is trained to detect morphing attacks.
SE-ResNeXt101-32x4d is a state-of-the-art image recognition model based on the ResNeXt101-32x4d model~\cite{resneXt} with the addition of the Squeeze-and-Excitation architecture~\cite{squeezenet}.
For all experiments a 5-fold stratified $k$-fold cross validation strategy is employed, thus preserving the class balance between morphed and bona fide images in each fold.
The model is fine-tuned on a collection of morphing attacks for 5 training epochs using exponential learning rate scheduler with differential learning rates in order to mitigate overfitting of the model.

\begin{figure}[t]
    \centering
    \begin{subfigure}{0.5\textwidth}
        \includegraphics[width=0.9\textwidth]{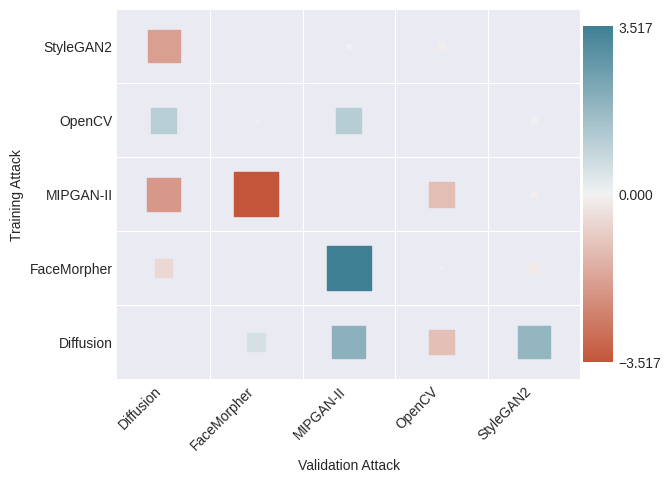}
        \caption{RSM on FRGC}
    \end{subfigure}
    \begin{subfigure}{0.5\textwidth}
        \includegraphics[width=.9\textwidth]{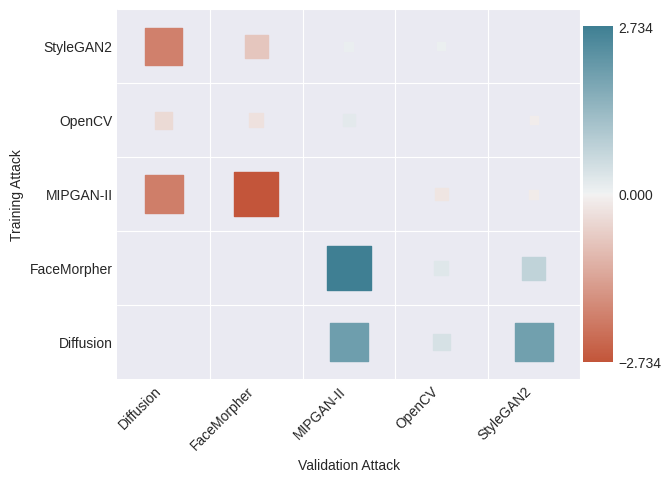}
        \caption{RSM on FERET}
    \end{subfigure}
    \begin{subfigure}{0.5\textwidth}
        \includegraphics[width=.9\textwidth]{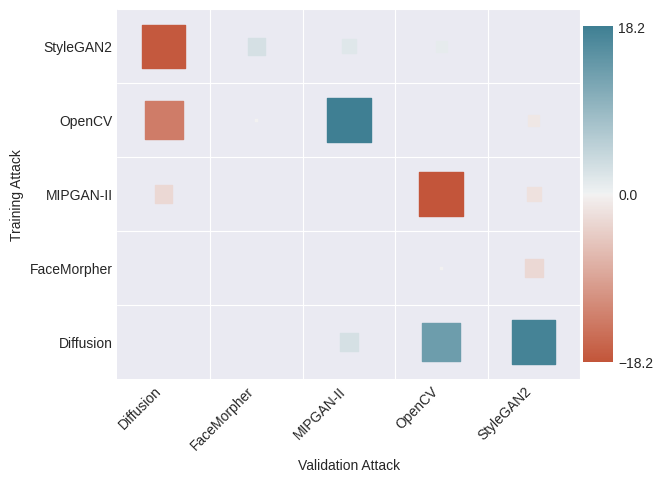}
        \caption{RSM on FRLL}
    \end{subfigure}
    \caption{Blue indicates higher strength and red indicates weaker strength.}
    \label{fig:rel-vggface}
\end{figure}

\subsubsection{Ablation Study}
To study the impact of a particular morphing attack on the ability of a MAD algorithm to detect morphing attacks, an ablation study was conducted where the SE-ResNeXt101-32x4d model was trained on all the morphing attacks except for one holdout.
\cref{tab:holdout} shows the validation accuracy of each morphing attack when different morphing attacks were withheld from the training process.
Due to the similar natures between the OpenCV and FaceMorpher attacks, the absence of one of these attacks does not greatly impact the validation accuracy.
Interestingly, the absence of MIPGAN-II does not significantly change the validation accuracy of the attacks; however, the omission of StyleGAN2 during training does decrease the performance of the StyleGAN2 during validation, despite the presence of MIPGAN-II.
Notably, the Diffusion attack is very difficult to detect as a novel attack, which can at least be partially attributed to its unique morph generation process in contrast with other morphing attacks.

\subsubsection{A Metric For Relative Strength}
In this section we introduce a new metric to measure the strength of one morphing attack relative to another.
We say a morph $\alpha$ is ``strong'' relative to a morph $\beta$ if the following conditions are satisfied:
\begin{enumerate}
    \item It is easy to detect $\beta$ when a detector is trained on $\alpha$, \ie, high transferability.
    \item It is hard to detect $\alpha$ when a detector is trained on $\beta$, \ie, low detectability.
\end{enumerate}
Additionally, the relative strength metric, $\Delta(\alpha||\beta)$, should be positive when $\alpha$ is stronger than $\beta$ and negative when $\alpha$ is weaker.
A relative strength of 0 would denote that the two morphing attacks are equally strong.

As some of the morphing attacks are not deterministic but probabilistic, we choose to represent a morphing attack $\alpha$ by the random variable $X^\alpha: \Omega \to \X$ such that $P(X^\alpha | x_a, x_b)$ denotes the distribution of morphs generated from images $x_a, x_b$.
Moreover, we suppose there exists a detector $f^\alpha: \X \to \{0,1\}$ trained to distinguish between bona fide presentations and morphed presentations generated by $\alpha$; wherein 0 denotes a bona fide presentation and 1 denotes a morphed presentation.
The transferability of a morphing attack $\alpha$ to $\beta$ is defined as the probability the detector $f^\alpha$ is able to detect the attack $\beta$ given the probability $f^\alpha$ detects $\alpha$, \ie, $T(\alpha, \beta) = P(f^\alpha(X^\beta) = 1 | f^\alpha(X^\alpha) = 1)$.
This metric can be represented as a ratio of expectations taken over the pairs of component bona fide images:
\begin{align}
T(\alpha&, \beta) = \frac{P(f^\alpha(X^\beta) = 1, f^\alpha(X^\alpha) = 1)}{P(f^\alpha(X^\alpha) = 1)}\nonumber\\
&= \frac{\ex_{x_a, x_b}[P(f^\alpha(X^\beta) = 1, f^\alpha(X^\alpha) = 1 \mid x_a, x_b)]}{\ex_{x_a, x_b}[P(f^\alpha(X^\alpha) = 1 \mid x_a, x_b)]}
\label{eq:transfer}
\end{align}
Let $\{x_i^\alpha\}_{i=1}^N$ denote a collection of $N$ samples drawn from $P(X^\alpha|x_a, x_b)$ such that $x_i^\alpha$ denotes the morph generated from $i$-th pair of bona fide identities $(a_i, b_i)$, and likewise for $\beta$.
Then the metric in~\cref{eq:transfer} can be closely approximated by
\begin{equation}
    T(\alpha, \beta) \approx \frac{\sum_{i=1}^N \Big [f^\alpha(x_i^\beta) =  1 \land f^\alpha(x_i^\alpha) = 1 \Big]}{\sum_{i=1}^{N} \Big[ f^\alpha(x_i^\alpha) = 1\Big]}
\end{equation}
\ie, the number of morphs from both $\alpha$ and $\beta$ detected over the number of morphs detected from $\alpha$.

The relative strength metric (RSM) from $\alpha$ to $\beta$ is defined as the log ratio of the transferability metrics between the two morphing attacks:
\begin{equation}
    \Delta(\alpha \| \beta) = \log \bigg(\frac{T(\alpha, \beta)}{T(\beta, \alpha)} \bigg)
\end{equation}
The log of the ratio is chosen such that the RSM takes positive values when $\alpha$ is ``stronger'' than $\beta$ and negative values when weaker---with a value of zero denoting equal strength.
Additionally, there is an antisymmetry such that $\Delta(\alpha\|\beta) = -\Delta(\beta\|\alpha)$.

In contrast to the ablation study, the SE-ResNeXt101-32x4d model is only trained on a single attack per $k$-fold.
The RSM is calculated between all attacks with the results shown in \cref{fig:rel-vggface}.
It can be observed that the RSM between the Landmark-based morphs and the RSM between the StyleGAN-based morphs is very small.
As these attacks have similar morph generation pipelines, it makes sense that the transferability between the attacks is near identical.
In general, the Landmark-based attacks seem to be stronger than the StyleGAN-based attacks, in particular the FaceMorpher attack.
The MIPGAN-II attack is generally weaker than the other attacks.
Overall, the Diffusion attack is the least detectable among the attacks along with generally being the strongest attack across the three datasets.

The results from \cref{fig:rel-vggface} corroborate with the results from \cref{tab:holdout}, demonstrating the difficulty in detecting Diffusion attacks.
From the perspective of training a MAD system, including samples from the FaceMorpher, StyleGAN, and Diffusion attacks would greatly increase the ability for the system to detect unknown attacks.
Additionally, \cref{tab:holdout} and \cref{fig:rel-vggface} demonstrates a particular vulnerability existing MAD systems may have towards the emerging Diffusion attack.

\begin{figure}[t]
    \centering
    \begin{subfigure}{0.24\textwidth}
        \includegraphics[width=.9\textwidth]{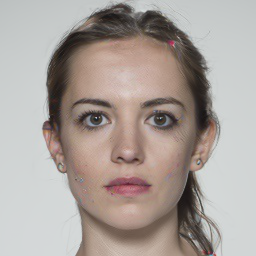}
        \caption{Variant A}
    \end{subfigure}%
    \begin{subfigure}{0.24\textwidth}
        \includegraphics[width=.9\textwidth]{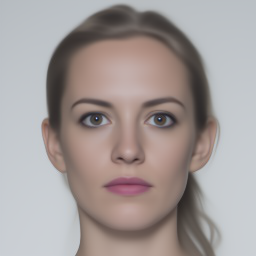}
        \caption{Variant B}
    \end{subfigure}
    \begin{subfigure}{0.24\textwidth}
        \includegraphics[width=.9\textwidth]{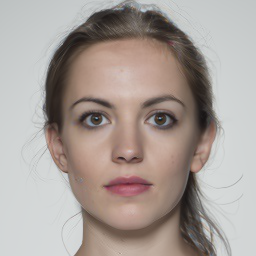}
        \caption{Variant C}
    \end{subfigure}%
    \begin{subfigure}{0.24\textwidth}
        \includegraphics[width=.9\textwidth]{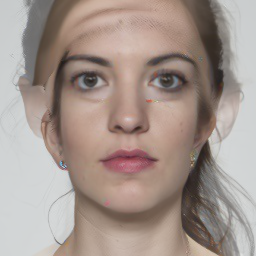}
        \caption{Variant D}
    \end{subfigure}
    \caption{Morphed image generated by different Diffusion attack variants on FRLL.}
    \label{fig:diff_variants}
\end{figure}

\begin{table*}[t]
    \centering
    \caption{MMPMR at FMR = 0.1\% across different configurations. Higher is better. \textdagger~indicates our default choices. \textdaggerdbl~the geometric mean.}
    \begin{adjustbox}{width=\linewidth}
    \begin{tabular}{lllrrrrrrrrrr}
        \toprule
        & & & \multicolumn{3}{c}{\textbf{FRLL}}
        & \multicolumn{3}{c}{\textbf{FRGC}}
        & \multicolumn{3}{c}{\textbf{FERET}}\\
        \cmidrule(lr){4-6}
        \cmidrule(lr){7-9}
        \cmidrule(lr){10-12}
        \textbf{Variant} & $\ell_\X$ & $\xi(x, y)$ &
        \textbf{FaceNet} & \textbf{VGGFace2} & \textbf{ArcFace} &
        \textbf{FaceNet} & \textbf{VGGFace2} & \textbf{ArcFace} &
        \textbf{FaceNet} & \textbf{VGGFace2} & \textbf{ArcFace} & \textbf{Mean\textsuperscript{\textdaggerdbl}}\\
        \midrule
        A&$\textrm{slerp}$ & $x, y \mapsto x$  &          \textbf{32.97} &        34.71 &          \textbf{88.27} &           \textbf{3.2}  &         \textbf{9.59} &          44.83 &            \textbf{7.17} &         \textbf{11.54} &           58.69 & \textbf{20.62}\\
        B&$\textrm{lerp}$ & $x, y \mapsto x$ &          10.81 &        11    &          68.94 &           1.17 &         2.17 &          38.85 &            2.33 &          4.69 &           35.41 & 8.79\\
        C\textsuperscript\textdagger &$\textrm{slerp}$ & $x, y \mapsto \frac{1}{2}(x + y)$ &          28.14 &        \textbf{35.37} &          88.09 &           2.68 &         8.47 &          \textbf{46.74} &            6.47 &         13.03 &           \textbf{59.75} & 19.80\\
        D& $\textrm{slerp}$ & $x, y \mapsto \textrm{OpenCV}(x, y)$ &           9.14 &         9.34 &          29.84 &           0    &         1.37 &           1.85 &            0.14 &          1.42 &            1.65 & 0.0\\
        \bottomrule
    \end{tabular}
    \end{adjustbox}
    \label{tab:mmpmr_configs}
\end{table*}

\subsection{Study of the Diffusion-based Morphing Process}

The diffusion morphing algorithm leverages both a stochastic and semantic representation of an image.
While the semantic representation contains many of the key ``identifying'' features, the stochastic representation contains many of the details necessary for high visual fidelity.
Due to the importance of the stochastic code for high fidelity, we investigated several methods for finding the morphed stochastic latent code, $\bfx_T^{(ab)}$.
The first variant, variant A, is the baseline implementation with $\ell_\Z$ using linear interpolation, $\ell_\X$ using spherical linear interpolation, and $\xi$ does not perform any ``pre-morphing''.
Conversely, in variant B the stochastic codes are interpolated via linear interpolation.
In variants C and D, instead of using the original image to calculate the stochastic code, the function $\xi$ is used to construct the ``pre-morph'' passed to the stochastic encoder.
Specifically, in variant C the two images are averaged pixel-wise and presented to the stochastic encoder; in contrast, in variant D the OpenCV morph is presented to the stochastic encoder. For each variant, we generated the same number of morphs across the three datasets as shown in Section \ref{42dataset}, resulting over 10,000 new morphs across four variants.

\begin{table}[t]
    \centering
    \caption{FID across different configurations. Lower is better. \textdagger~indicates our default choices.}
    \begin{adjustbox}{width=\linewidth}
    \begin{tabular}{lllrrr}
        \toprule
        \textbf{Variant} & $\ell_\X$ & $\xi(x, y)$ & \textbf{FRLL} & \textbf{FRGC} & \textbf{FERET}\\
        \midrule
        A&$\textrm{slerp}$ & $x, y \mapsto x$ & 48.13 & \textbf{52.97} & 55.66\\
        B&$\textrm{lerp}$ & $x, y \mapsto x$ & 82.05 & 119.33 & 97.75\\
        C\textsuperscript\textdagger&$\textrm{slerp}$ & $x, y \mapsto \frac{1}{2}(x + y)$ & \textbf{42.63} & 64.16 & \textbf{50.45}\\
        D&$\textrm{slerp}$ & $x, y \mapsto \textrm{OpenCV}(x, y)$ & 93.85 & 84.51 & 108.49\\
        \bottomrule
    \end{tabular}
    \end{adjustbox}
    \label{tab:fid_diff}
\end{table}

In \cref{tab:fid_diff} the FID is calculated between the generated morphs and the bona fide samples for each particular dataset.
Variant C generally presents the lowest FID score, closely followed by variant A.
Both variants B and D exhibit clear degradation in performance when compared to variants A and C.
Furthermore, the FID score seems to correlate well with human assessment of the generated samples, see \cref{fig:diff_variants}.
Noticeably, the linear interpolation in variant B results in an overly smoothed face and generally darker image, greatly degrading visual fidelity.
Variant D has prominent visual artifacts, similar to the artifacts found in the OpenCV morphs.
Moreover, the poor performance seems to be aided by an issue of differing alignment strategies between the OpenCV and diffusion pipeline.

Notably, variant C often removes many of the high frequency artifacts found in variant A.
This is likely due to the difficulty in smoothly interpolating between points in the stochastic latent space, in contrast with the semantic latent space.
As such, variant C, which performs a pixel-wise average of the two source images before using the stochastic encoder, seems to greatly improve the ability to smoothly interpolate between different stochastic representations.
This appears to be the primary reason variant C has a generally lower FID when compared to variant A.
Both \cref{fig:diff_variants} and \cref{tab:fid_diff} demonstrate the large importance that the stochastic code plays in creating high fidelity morphed images.
Due to the high fidelity exhibited by variant C,
this particular diffusion process was used in evaluation against other morphing attacks.

The MMPMR metric is calculated for each variant in \cref{tab:mmpmr_configs}.
Variant A is slightly stronger than variant C, with variants B and D falling far behind, likely due to the high number of visual distortions.
These results stand in contrast to the assessment of visual fidelity wherein variant C outperforms variant A.
This, again, illustrates a trade-off between visual fidelity and ability to fool the FR system; however, in this case the trade-off effectiveness against the FR system is relatively small in comparison to the gains in visual fidelity.
Due to its excellent visual fidelity and strong MMPMR results, variant C was still chosen to be the default configuration for the Diffusion attack.

\section{Conclusions}

By addressing some of the key limitations of prior deep-learning based morphing attacks, namely, the trade-off between visual fidelity and effectiveness against FR systems, we have proposed a novel morphing attack using Diffusion-based methods for the generative process.
The proposed attack consistently generates realistic morphed images with high visual fidelity, while also being able to strongly threaten FR systems.
To evaluate the attack potential of the proposed method, we evaluated the vulnerability of three FR systems over three distinct datasets 
across four variants of the Diffusion morph, with the strongest variant achieving state-of-the-art performance.
We also studied the impact of the pre-processing pipeline on the vulnerability of an FR system to morphing attacks.
A novel metric to assess the strength of one morphing attack relative to another has been introduced.
Moreover, the proposed attack was evaluated by its detection performance against a state-of-the-art MAD system.
The Diffusion attack was shown to be very difficult to detect if not specifically trained against, showing the proposed attack can greatly threaten existing FR systems.
Overall, the images generated by the Diffusion attack possess high visual fidelity, can deceive state-of-the-art FR systems, and are difficult for MAD mechanisms to detect. 
It is our belief that developing a stronger face morphing attack is essential to the design and testing of stronger MAD methods.
As far as future work is concerned, the proposed method could be extended to higher resolutions using diffusion-based super resolution methods or latent diffusion methods.

\ifCLASSOPTIONcompsoc
  \section*{Acknowledgments}
\else
  \section*{Acknowledgment}
\fi
This material is based upon work supported by the Center for Identification Technology Research and National Science Foundation under Grant \#1650503.

\ifCLASSOPTIONcaptionsoff
  \newpage
\fi
\bibliographystyle{IEEEtran}
\bibliography{bib}

\begin{thebibliography}{10}
\providecommand{\url}[1]{#1}
\csname url@samestyle\endcsname
\providecommand{\newblock}{\relax}
\providecommand{\bibinfo}[2]{#2}
\providecommand{\BIBentrySTDinterwordspacing}{\spaceskip=0pt\relax}
\providecommand{\BIBentryALTinterwordstretchfactor}{4}
\providecommand{\BIBentryALTinterwordspacing}{\spaceskip=\fontdimen2\font plus
\BIBentryALTinterwordstretchfactor\fontdimen3\font minus
  \fontdimen4\font\relax}
\providecommand{\BIBforeignlanguage}[2]{{%
\expandafter\ifx\csname l@#1\endcsname\relax
\typeout{** WARNING: IEEEtran.bst: No hyphenation pattern has been}%
\typeout{** loaded for the language `#1'. Using the pattern for}%
\typeout{** the default language instead.}%
\else
\language=\csname l@#1\endcsname
\fi
#2}}
\providecommand{\BIBdecl}{\relax}
\BIBdecl

\bibitem{frs-rates}
L.~J. {Spreeuwers}, A.~J. {Hendrikse}, and K.~J. {Gerritsen}, ``Evaluation of
  automatic face recognition for automatic border control on actual data
  recorded of travellers at schiphol airport,'' in \emph{Proc. of the Int'l
  Conf. of Biometrics Special Interest Group (BIOSIG)}, 2012, pp. 1--6.

\bibitem{Ferrara2016}
\BIBentryALTinterwordspacing
M.~Ferrara, A.~Franco, and D.~Maltoni, \emph{On the Effects of Image
  Alterations on Face Recognition Accuracy}.\hskip 1em plus 0.5em minus
  0.4em\relax Cham: Springer International Publishing, 2016, pp. 195--222.
  [Online]. Available: \url{https://doi.org/10.1007/978-3-319-28501-6_9}
\BIBentrySTDinterwordspacing

\bibitem{fraud_id}
\BIBentryALTinterwordspacing
D.~J. Robertson, R.~S.~S. Kramer, and A.~M. Burton, ``Fraudulent id using face
  morphs: Experiments on human and automatic recognition,'' \emph{PLOS ONE},
  vol.~12, no.~3, pp. 1--12, 03 2017. [Online]. Available:
  \url{https://doi.org/10.1371/journal.pone.0173319}
\BIBentrySTDinterwordspacing

\bibitem{morphed_first}
R.~{Raghavendra}, K.~B. {Raja}, and C.~{Busch}, ``Detecting morphed face
  images,'' in \emph{IEEE 8th Int'l Conf. on Biometrics Theory, Applications
  and Systems (BTAS)}, 2016, pp. 1--7.

\bibitem{multi_image_attacks}
\BIBentryALTinterwordspacing
J.~T.~A. Andrews, T.~Tanay, and L.~D. Griffin, ``Multiple-identity image
  attacks against face-based identity verification,'' \emph{CoRR}, vol.
  abs/1906.08507, 2019. [Online]. Available:
  \url{http://arxiv.org/abs/1906.08507}
\BIBentrySTDinterwordspacing

\bibitem{Raghavendra2017FaceMV}
R.~Raghavendra, K.~Raja, S.~Venkatesh, and C.~Busch, ``Face morphing versus
  face averaging: Vulnerability and detection,'' \emph{IEEE Int'l Joint Conf.
  on Biometrics (IJCB)}, pp. 555--563, 2017.

\bibitem{germany_morph_fraud}
C.~Burt, ``Face morphing threat to biometric identity credentials’
  trustworthiness a growing problem,'' \emph{BiometricUpdate.com}.

\bibitem{mor-face-det-deep-resid}
S.~{Venkatesh}, R.~{Ramachandra}, K.~{Raja}, L.~{Spreeuwers}, R.~{Veldhuis},
  and C.~{Busch}, ``Morphed face detection based on deep color residual
  noise,'' in \emph{the 9th Int'l Conf. on Image Pr. Theory, Tools and
  Applications (IPTA)}, 2019, pp. 1--6.

\bibitem{on-vuln}
U.~{Scherhag}, R.~{Raghavendra}, K.~B. {Raja}, M.~{Gomez-Barrero},
  C.~{Rathgeb}, and C.~{Busch}, ``On the vulnerability of face recognition
  systems towards morphed face attacks,'' in \emph{2017 5th International
  Workshop on Biometrics and Forensics (IWBF)}, 2017, pp. 1--6.

\bibitem{prnu}
U.~{Scherhag}, L.~{Debiasi}, C.~{Rathgeb}, C.~{Busch}, and A.~{Uhl},
  ``Detection of face morphing attacks based on prnu analysis,'' \emph{IEEE
  Transactions on Biometrics, Behavior, and Identity Science}, vol.~1, no.~4,
  pp. 302--317, 2019.

\bibitem{Blasingame2021LeveragingAL}
Z.~Blasingame and C.~Liu, ``Leveraging adversarial learning for the detection
  of morphing attacks,'' \emph{2021 IEEE International Joint Conference on
  Biometrics (IJCB)}, pp. 1--8, 2021.

\bibitem{survery}
S.~{Venkatesh}, R.~{Ramachandra}, K.~{Raja}, and C.~{Busch}, ``{Face Morphing
  Attack Generation \& Detection: A Comprehensive Survey},'' \emph{arXiv
  e-prints arXiv:2011.02045}, Nov. 2020.

\bibitem{Ojala1996ACS}
T.~Ojala, M.~Pietik{\"a}inen, and D.~Harwood, ``A comparative study of texture
  measures with classification based on featured distributions,'' \emph{Pattern
  Recognit.}, vol.~29, pp. 51--59, 1996.

\bibitem{bsif}
J.~{Kannala} and E.~{Rahtu}, ``Bsif: Binarized statistical image features,'' in
  \emph{Proceedings of the 21st Int'l Conf. on Pattern Recognition (ICPR2012)},
  2012, pp. 1363--1366.

\bibitem{vgg19-is-best}
C.~Seibold, W.~Samek, A.~Hilsmann, and P.~Eisert, ``Detection of face morphing
  attacks by deep learning,'' in \emph{Digital Forensics and Watermarking},
  C.~Kraetzer, Y.-Q. Shi, J.~Dittmann, and H.~J. Kim, Eds., 2017, pp. 107--120.

\bibitem{deep-best}
U.~{Scherhag}, C.~{Rathgeb}, J.~{Merkle}, and C.~{Busch}, ``Deep face
  representations for differential morphing attack detection,'' \emph{IEEE
  Transactions on Information Forensics and Security}, vol.~15, pp. 3625--3639,
  2020.

\bibitem{NISTFRVT}
M.~Ngan, P.~Grother, K.~Hanaoka, and J.~Kuo, ``\BIBforeignlanguage{en}{Face
  recognition vendor test (frvt) part 4: Morph - performance of automated face
  morph detection},'' 2020-03-06 2020.

\bibitem{sebastian_morphs}
E.~Sarkar, P.~Korshunov, L.~Colbois, and S.~Marcel, ``Vulnerability analysis of
  face morphing attacks from landmarks and generative adversarial networks,''
  \emph{ArXiv}, vol. abs/2012.05344, 2020.

\bibitem{nasser_wavelet}
K.~O'Haire, S.~Soleymani, B.~Chaudhary, P.~Aghdaie, J.~Dawson, and N.~M.
  Nasrabadi, ``Adversarially perturbed wavelet-based morphed face generation,''
  in \emph{2021 16th IEEE International Conference on Automatic Face and
  Gesture Recognition (FG 2021)}, 2021, pp. 01--05.

\bibitem{morgan}
N.~{Damer}, A.~M. {Saladié}, A.~{Braun}, and A.~{Kuijper}, ``Morgan:
  Recognition vulnerability and attack detectability of face morphing attacks
  created by generative adversarial network,'' in \emph{2018 IEEE 9th Int'l
  Conf. on Biometrics Theory, Applications and Systems (BTAS)}, 2018, pp.
  1--10.

\bibitem{can_gan_beat_landmark}
S.~Venkatesh, H.~Zhang, R.~Ramachandra, K.~Raja, N.~Damer, and C.~Busch, ``Can
  gan generated morphs threaten face recognition systems equally as landmark
  based morphs? - vulnerability and detection,'' in \emph{2020 8th
  International Workshop on Biometrics and Forensics (IWBF)}, 2020, pp. 1--6.

\bibitem{stylegan2}
T.~{Karras}, S.~{Laine}, M.~{Aittala}, J.~{Hellsten}, J.~{Lehtinen}, and
  T.~{Aila}, ``Analyzing and improving the image quality of stylegan,'' in
  \emph{2020 IEEE/CVF Conference on Computer Vision and Pattern Recognition
  (CVPR)}, 2020, pp. 8107--8116.

\bibitem{mipgan}
H.~Zhang, S.~Venkatesh, R.~Ramachandra, K.~Raja, N.~Damer, and C.~Busch,
  ``Mipgan—generating strong and high quality morphing attacks using identity
  prior driven gan,'' \emph{IEEE Transactions on Biometrics, Behavior, and
  Identity Science}, vol.~3, no.~3, pp. 365--383, 2021.

\bibitem{diff_beat_gan}
\BIBentryALTinterwordspacing
P.~Dhariwal and A.~Nichol, ``Diffusion models beat gans on image synthesis,''
  in \emph{Advances in Neural Information Processing Systems}, M.~Ranzato,
  A.~Beygelzimer, Y.~Dauphin, P.~Liang, and J.~W. Vaughan, Eds., vol.~34.\hskip
  1em plus 0.5em minus 0.4em\relax Curran Associates, Inc., 2021, pp.
  8780--8794. [Online]. Available:
  \url{https://proceedings.neurips.cc/paper/2021/file/49ad23d1ec9fa4bd8d77d02681df5cfa-Paper.pdf}
\BIBentrySTDinterwordspacing

\bibitem{frll}
L.~DeBruine and B.~Jones, ``Face research lab london set.''

\bibitem{venkatesh2020gan}
S.~Venkatesh, H.~Zhang, R.~Ramachandra, K.~Raja, N.~Damer, and C.~Busch, ``Can
  gan generated morphs threaten face recognition systems equally as landmark
  based morphs? -- vulnerability and detection,'' 2020.

\bibitem{sebastian_gan_threaten}
E.~Sarkar, P.~Korshunov, L.~Colbois, and S.~Marcel, ``Are gan-based morphs
  threatening face recognition?'' in \emph{ICASSP 2022 - 2022 IEEE
  International Conference on Acoustics, Speech and Signal Processing
  (ICASSP)}, 2022, pp. 2959--2963.

\bibitem{facemorpher}
A.~Quek, ``{Facemorpher},'' \url{https://github.com/alyssaq/face_morpher},
  2019.

\bibitem{STASM}
S.~{Milborrow} and F.~{Nicolls}, ``Active shape models with sift descriptors
  and mars,'' in \emph{2014 International Conference on Computer Vision Theory
  and Applications (VISAPP)}, vol.~2, 2014, pp. 380--387.

\bibitem{multe-scale-block-fusion}
U.~Scherhag, J.~Kunze, C.~Rathgeb, and C.~Busch, ``Face morph detection for
  unknown morphing algorithms and image sources: a multi-scale block local
  binary pattern fusion approach,'' \emph{IET Biometrics}, vol.~9, pp.
  278--289, 11 2020.

\bibitem{dlib}
D.~E. King, ``Dlib-ml: A machine learning toolkit,'' \emph{J. Mach. Learn.
  Res.}, vol.~10, p. 1755–1758, Dec. 2009.

\bibitem{sebastian_on_detection_of_ma_gan}
L.~Colbois and S.~Marcel, ``On the detection of morphing attacks generated by
  gans,'' in \emph{2022 International Conference of the Biometrics Special
  Interest Group (BIOSIG)}, 2022, pp. 1--5.

\bibitem{Roich2021PivotalTF}
D.~Roich, R.~Mokady, A.~H. Bermano, and D.~Cohen-Or, ``Pivotal tuning for
  latent-based editing of real images,'' \emph{ACM Transactions on Graphics
  (TOG)}, 2021.

\bibitem{stylegan}
T.~{Karras}, S.~{Laine}, and T.~{Aila}, ``A style-based generator architecture
  for generative adversarial networks,'' in \emph{2019 IEEE/CVF Conference on
  Computer Vision and Pattern Recognition (CVPR)}, 2019, pp. 4396--4405.

\bibitem{biggandeep}
\BIBentryALTinterwordspacing
A.~Brock, J.~Donahue, and K.~Simonyan, ``Large scale {GAN} training for high
  fidelity natural image synthesis,'' in \emph{International Conference on
  Learning Representations}, 2019. [Online]. Available:
  \url{https://openreview.net/forum?id=B1xsqj09Fm}
\BIBentrySTDinterwordspacing

\bibitem{ddpm}
\BIBentryALTinterwordspacing
J.~Ho, A.~Jain, and P.~Abbeel, ``Denoising diffusion probabilistic models,'' in
  \emph{Advances in Neural Information Processing Systems}, H.~Larochelle,
  M.~Ranzato, R.~Hadsell, M.~Balcan, and H.~Lin, Eds., vol.~33.\hskip 1em plus
  0.5em minus 0.4em\relax Curran Associates, Inc., 2020, pp. 6840--6851.
  [Online]. Available:
  \url{https://proceedings.neurips.cc/paper/2020/file/4c5bcfec8584af0d967f1ab10179ca4b-Paper.pdf}
\BIBentrySTDinterwordspacing

\bibitem{unet}
O.~Ronneberger, P.~Fischer, and T.~Brox, ``U-net: Convolutional networks for
  biomedical image segmentation,'' in \emph{Medical Image Computing and
  Computer-Assisted Intervention -- MICCAI 2015}, N.~Navab, J.~Hornegger, W.~M.
  Wells, and A.~F. Frangi, Eds.\hskip 1em plus 0.5em minus 0.4em\relax Cham:
  Springer International Publishing, 2015, pp. 234--241.

\bibitem{song2021denoising}
\BIBentryALTinterwordspacing
J.~Song, C.~Meng, and S.~Ermon, ``Denoising diffusion implicit models,'' in
  \emph{International Conference on Learning Representations}, 2021. [Online].
  Available: \url{https://openreview.net/forum?id=St1giarCHLP}
\BIBentrySTDinterwordspacing

\bibitem{diffae}
K.~Preechakul, N.~Chatthee, S.~Wizadwongsa, and S.~Suwajanakorn, ``Diffusion
  autoencoders: Toward a meaningful and decodable representation,'' in
  \emph{Proceedings of the IEEE/CVF Conference on Computer Vision and Pattern
  Recognition (CVPR)}, June 2022, pp. 10\,619--10\,629.

\bibitem{pytorch}
A.~Paszke, S.~Gross, S.~Chintala, G.~Chanan, E.~Yang, Z.~DeVito, Z.~Lin,
  A.~Desmaison, L.~Antiga, and A.~Lerer, ``Automatic differentiation in
  pytorch,'' 2017.

\bibitem{vggface2}
Q.~Cao, L.~Shen, W.~Xie, O.~M. Parkhi, and A.~Zisserman, ``Vggface2: A dataset
  for recognising faces across pose and age,'' in \emph{2018 13th IEEE
  International Conference on Automatic Face \& Gesture Recognition (FG 2018)},
  2018, pp. 67--74.

\bibitem{facenet}
F.~{Schroff}, D.~{Kalenichenko}, and J.~{Philbin}, ``Facenet: A unified
  embedding for face recognition and clustering,'' in \emph{2015 IEEE
  Conference on Computer Vision and Pattern Recognition (CVPR)}, 2015, pp.
  815--823.

\bibitem{deng2019arcface}
J.~Deng, J.~Guo, N.~Xue, and S.~Zafeiriou, ``Arcface: Additive angular margin
  loss for deep face recognition,'' in \emph{Proceedings of the IEEE Conference
  on Computer Vision and Pattern Recognition}, 2019, pp. 4690--4699.

\bibitem{vggface}
O.~M. Parkhi, A.~Vedaldi, and A.~Zisserman, ``Deep face recognition,'' in
  \emph{BMVC}, 2015.

\bibitem{squeezenet}
J.~Hu, L.~Shen, and G.~Sun, ``Squeeze-and-excitation networks,'' in \emph{2018
  IEEE/CVF Conference on Computer Vision and Pattern Recognition}, 2018, pp.
  7132--7141.

\bibitem{iresnet}
I.~C. Duta, L.~Liu, F.~Zhu, and L.~Shao, ``Improved residual networks for image
  and video recognition,'' \emph{arXiv preprint arXiv:2004.04989}, 2020.

\bibitem{glint360k}
X.~An, X.~Zhu, Y.~Gao, Y.~Xiao, Y.~Zhao, Z.~Feng, L.~Wu, B.~Qin, M.~Zhang,
  D.~Zhang, and Y.~Fu, ``Partial fc: Training 10 million identities on a single
  machine,'' in \emph{2021 IEEE/CVF International Conference on Computer Vision
  Workshops (ICCVW)}, 2021, pp. 1445--1449.

\bibitem{mtcnn}
K.~Zhang, Z.~Zhang, Z.~Li, and Y.~Qiao, ``Joint face detection and alignment
  using multitask cascaded convolutional networks,'' \emph{IEEE Signal
  Processing Letters}, vol.~23, no.~10, pp. 1499--1503, 2016.

\bibitem{feret}
P.~Phillips, H.~Wechsler, J.~Huang, and P.~J. Rauss, ``The feret database and
  evaluation procedure for face-recognition algorithms,'' \emph{Image Vis.
  Comput.}, vol.~16, pp. 295--306, 1998.

\bibitem{frgc}
P.~Phillips, P.~Flynn, T.~Scruggs, K.~Bowyer, J.~Chang, K.~Hoffman, J.~Marques,
  J.~Min, and W.~Worek, ``Overview of the face recognition grand challenge,''
  in \emph{2005 IEEE Computer Society Conference on Computer Vision and Pattern
  Recognition (CVPR'05)}, vol.~1, 2005, pp. 947--954 vol. 1.

\bibitem{FID}
M.~Lucic, K.~Kurach, M.~Michalski, O.~Bousquet, and S.~Gelly, ``Are gans
  created equal? a large-scale study,'' in \emph{Proceedings of the 32nd
  International Conference on Neural Information Processing Systems}, ser.
  NIPS'18.\hskip 1em plus 0.5em minus 0.4em\relax Red Hook, NY, USA: Curran
  Associates Inc., 2018, p. 698–707.

\bibitem{imagenet}
J.~Deng, W.~Dong, R.~Socher, L.-J. Li, K.~Li, and L.~Fei-Fei, ``Imagenet: A
  large-scale hierarchical image database,'' in \emph{2009 IEEE Conference on
  Computer Vision and Pattern Recognition}, 2009, pp. 248--255.

\bibitem{Seitzer2020FID}
M.~Seitzer, ``{pytorch-fid: FID Score for PyTorch},''
  \url{https://github.com/mseitzer/pytorch-fid}, August 2020, version 0.2.1.

\bibitem{frontex}
FRONTEX, ``Best practice technical guidelines for automated border control abc
  systems,'' Frontex, the European Border and Coast Guard Agency, Plac
  Europejski 6, 00-844 Warsaw, Poland, Tech. Rep., 2015.

\bibitem{mmpmr}
U.~Scherhag, A.~Nautsch, C.~Rathgeb, M.~Gomez-Barrero, R.~N.~J. Veldhuis,
  L.~Spreeuwers, M.~Schils, D.~Maltoni, P.~Grother, S.~Marcel, R.~Breithaupt,
  R.~Ramachandra, and C.~Busch, ``Biometric systems under morphing attacks:
  Assessment of morphing techniques and vulnerability reporting,'' in
  \emph{2017 International Conference of the Biometrics Special Interest Group
  (BIOSIG)}, 2017, pp. 1--7.

\bibitem{stylegan3}
T.~Karras, M.~Aittala, S.~Laine, E.~H\"ark\"onen, J.~Hellsten, J.~Lehtinen, and
  T.~Aila, ``Alias-free generative adversarial networks,'' in \emph{Proc.
  NeurIPS}, 2021.

\bibitem{resneXt}
S.~Xie, R.~B. Girshick, P.~Doll{\'a}r, Z.~Tu, and K.~He, ``Aggregated residual
  transformations for deep neural networks,'' \emph{2017 IEEE Conference on
  Computer Vision and Pattern Recognition (CVPR)}, pp. 5987--5995, 2017.

\end{thebibliography}

\begin{IEEEbiography}[{\includegraphics[angle=0,width=1in,height=1.25in,clip,keepaspectratio]{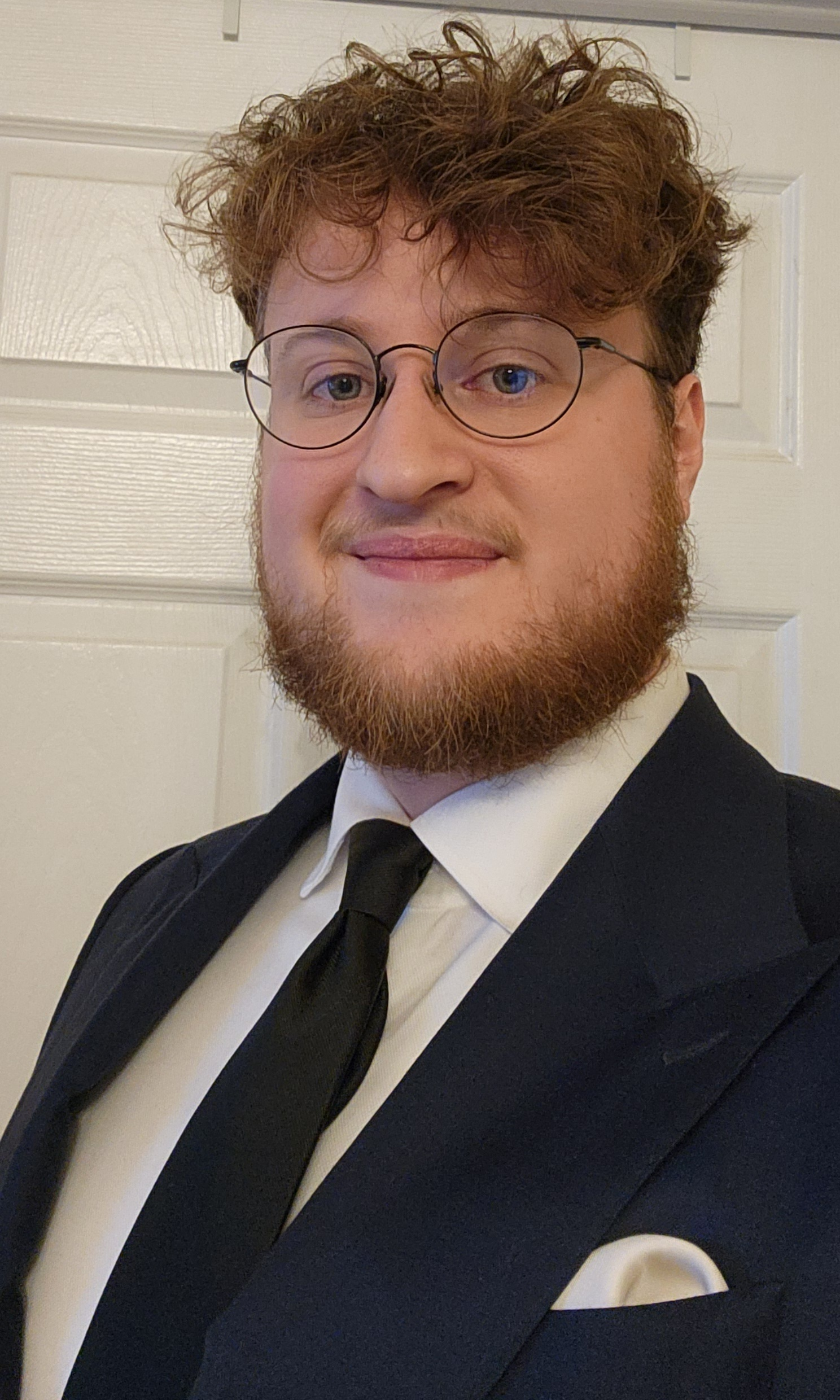}}]{Zander Blasingame}
received the B.Sc. degree in Computer Engineering from Clarkson University in 2018. He is currently pursuing a Ph.D. degree in Electrical and Computer Engineering at Clarkson University, Potsdam, New York. His research interests in biometrics include face morphing, digital face generation, latent representations of faces, and latent manipulations of faces which are mainly funded through the Center for Identification Technology Research (CITeR), a National Science Foundation Industry/University Cooperative Research Center.
\end{IEEEbiography}
\begin{IEEEbiography}[{\includegraphics[width=1in,height=1.25in,clip,keepaspectratio]{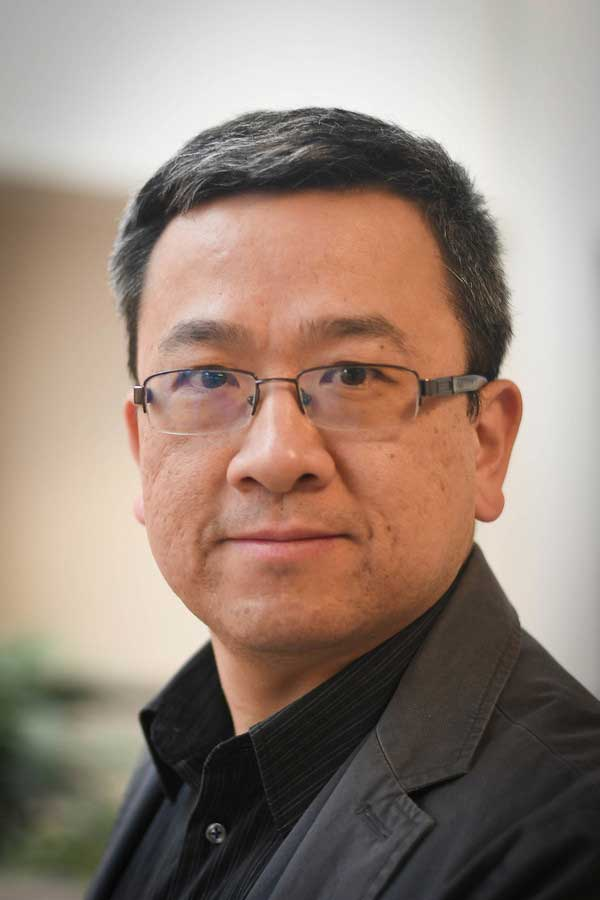}}]{Chen Liu}
(Senior Member, IEEE) received the Ph.D. degree in Electrical and Computer Engineering from the University of California, Irvine (UCI), Irvine, CA, USA, in 2008. He is currently an Associate Professor with the Department of Electrical and Computer Engineering at Clarkson University, Potsdam, NY, USA. His research interests in biometrics include face in video recognition (FiVR), face quality evaluation, face morphing and real-time FiVR via hardware acceleration, which are mainly funded through the Center for Identification Technology Research (CITeR), a National Science Foundation Industry/University Cooperative Research Center. He is also a Senior Member of ACM.
\end{IEEEbiography}

\end{document}